\documentclass[journal]{IEEEtran}

\usepackage[utf8]{inputenc}
\usepackage{tikz}
\usepackage{subcaption}
\usepackage{xcolor}
\usepackage{amssymb}
\usepackage{algpseudocode}
\usetikzlibrary {arrows,calc,bayesnet,shapes,backgrounds,decorations.pathreplacing,shadows.blur,positioning}

\tikzset{
	-|-/.style={
		to path={
			(\tikztostart) -| ($(\tikztostart)!#1!(\tikztotarget)$) |- (\tikztotarget)
			\tikztonodes
		}
	},
	-|-/.default=0.5,
	|-|/.style={
		to path={
			(\tikztostart) |- ($(\tikztostart)!#1!(\tikztotarget)$) -| (\tikztotarget)
			\tikztonodes
		}
	},
	|-|/.default=0.5,
}

\begin{document}

\title{Redefining Neural Architecture Search of Heterogeneous Multi-Network Models by Characterizing Variation Operators and Model Components}
%
%
%

\author{Unai Garciarena,
        Roberto Santana
        and~Alexander Mendiburu
\thanks{U. Garciarena and R. Santana are with the Department of Computer Science and Artificial Intelligence, University of the Basque Country UPV/EHU, 20018, Donostia, Spain.}
\thanks{A. Mendiburu is with the Department of Computer Architecture and Technology, University of the Basque Country UPV/EHU, 20018, Donostia, Spain.}
\thanks{email: unai.garciarena@ehu.eus (corresponding author), roberto.santana@ehu.eus, alexander.mendiburu@ehu.eus}
\thanks{© 2022 IEEE. Personal use of this material is permitted. Permission from IEEE must be obtained for all other uses, including reprinting/republishing this material for advertising or promotional purposes, collecting new collected works for resale or redistribution to servers or lists, or reuse of any copyrighted component of this work in other works.}
\thanks{This work has been submitted to the IEEE for possible publication. Copyright may be transferred without notice, after which this version may no longer be accessible.}}

%
%

\markboth{Journal of \LaTeX\ Class Files,~Vol.~14, No.~8, May~2021}%
{Redefining Neural Architecture Search of Heterogeneous Multi-Network Models by Characterizing Variation Operators and Model Components}
%



\maketitle

\begin{abstract}
With neural architecture search methods gaining ground on manually designed deep neural networks -even more rapidly as model sophistication escalates-, the research trend shifts towards arranging different and often increasingly complex neural architecture search spaces. In this conjuncture, delineating algorithms which can efficiently explore these search spaces can result in a significant improvement over currently used methods, which, in general, randomly select the structural variation operator, hoping for a performance gain. In this paper, we investigate the effect of different variation operators in a complex domain, that of multi-network heterogeneous neural models. These models have an extensive and complex search space of structures as they require multiple sub-networks within the general model in order to answer to different output types. From that investigation, we extract a set of general guidelines, whose application is not limited to that particular type of model, and are useful to determine the direction in which an architecture optimization method could find the largest improvement. To deduce the set of guidelines, we characterize both the variation operators, according to their effect on the complexity and performance of the model; and the models, relying on diverse metrics which estimate the quality of the different parts composing it.
\end{abstract}


%
\IEEEpeerreviewmaketitle

\section{Introduction}

Deep neural networks (DNN) are considered to be black-box models due to their opaqueness when it comes to the iterpretability of their operation mechanics. However, this fact has not deterred researchers from studying their application to many research fields, attracted by their impressive performance in several different domains \cite{goodfellow_deep_2016}. This phenomenon is particularly noticeable in image-related tasks: classification \cite{krizhevsky_imagenet_2012,szegedy_rethinking_2016}, captioning \cite{vinyals_show_2015}, or segmentation \cite{badrinarayanan_segnet_2017}; albeit other areas such as natural language processing \cite{jozefowicz_exploring_2016} or data generation \cite{goodfellow_generative_2014} have also been benefited.

Their adaptability to different problem specifications has resulted in an enormous spike in DNN research in the last decade. Initially little or no attention was paid to the DNN structure (a term that we use to comprise the architecture of a DNN as well as other hyperparameters, such as the training specification), and default or baseline designs were used. However, several researchers noticed that finding the right structure proved to be essential for obtaining a good performing DNN for the particular problem at hand \cite{krizhevsky_imagenet_2012,szegedy_going_2015,szegedy_rethinking_2016}. This fact promoted the usage of hand-made designs, guided by expert knowledge. This approach, to this day, still enjoys a large share of popularity \cite{sandler_mobilenetv2_2018,szegedy_inception-v4_2017}. However, the popularization of DNNs and their application to more and more challenging problems has proportionally escalated the necessity for increasingly complex DNN structures, to the point where designing them has become a process too time-consuming and difficult to be carried out by hand.

In order to overcome this issue, several approaches aiming at neural architecture search (NAS) have been proposed in the last years, their main goal being to extricate humans from the duty of manually designing neural models, as well as being able to obtain structures \cite{elsken_neural_2019,liang_evolutionary_2018,zoph_learning_2018} which fit a given problem exceptionally well. These techniques have evolved from relying on modest operators whose scope merely enabled them to perform small changes \cite{stanley_efficient_2002} -such as modifying one parameter or connection-, to directly adding complete neural cells that could be considered DNNs themselves \cite{liang_evolutionary_2018}. Some representative examples of NAS techniques employ reinforcement learning \cite{francois-lavet_introduction_2018}, evolutionary algorithms (EA) \cite{miikkulainen_evolving_2019}, or local search methods \cite{elsken_simple_2017}.

The lack of efficiency is a flaw often held against NAS algorithms, as, commonly, assessing the quality of a DNN structure involves weight optimization procedures, which tend to be rather costly \cite{white_local_2020}. Hence, our main concern when conceiving NAS algorithms is sharpening the usage of every component of the search algorithm so that we make the most out of every evaluation. The efficiency of NAS algorithms is largely dependent on the effectiveness of the operators they employ. Moreover, that effectiveness could fluctuate depending on when or where it is applied, i.e., a productive operator can result useless if applied in the wrong circumstances. 

In this context, we identify a large potential for improvement in these NAS methods, as most of them apply modification operators randomly, hoping for an improvement in the resulting model. Developing criteria to select the most suitable choice from a pool of operators, and the part of the model to be the target of that modification, would improve the efficiency of these NAS algorithms.

Structural specialization is not the only route in which DNNs have made progress. As a matter of fact, research areas which establish their base on completely antipodal intuition have also received their fair share of attention. A good example of this is multi-task learning (MTL) \cite{caruana_multitask_1997}, a learning paradigm which employs one single model for predicting similar targets from pieces of data which follow similar distributions. This approach has proven to have more benefits apart from the obvious parameter reduction in the DNN, as it has been demonstrated that the multiple tasks \textit{coexisting} in a single model act as regularizers of each other by introducing an inductive bias, since the model cannot focus on learning a single task. Furthermore, it has been theoretically proven \cite{liu_algorithm-dependent_2016} that the value of the loss function of a task within an MTL framework tends to the same value that it would have obtained had it been learned separately, as the number of training observations increases. Although not as popular as the ones introduced at the beginning of this section, problems in which multiple outputs have to be addressed simultaneously also exist, and DNNs (of elevated complexity) have recently been the model of choice to approach them \cite{garciarena_towards_2021,li_person_2017,zheng_joint_2019}. This does nothing but further intensify the need for automatic DNN structure designing algorithms. Furthermore, while extending the complexity of the networks results in an increase in the time required to researchers for developing new structures that outperform the previous ones, the progress made regarding the hardware in which DNN training can be parallelized pushes the time elapsed by NAS techniques in the opposite direction \cite{rupp_cpu_2016}. One particular variant of MTL, heterogeneous MTL (HMTL) was proposed in \cite{garciarena_towards_2021}. Unlike traditional MTL, this new research line focuses on developing models which are capable of dealing with several problems simultaneously (potentially of different nature, e.g., classification, regression and data generation) and in a collaborative way. This is achieved by designing a single and complex DNN susceptible of being incrementally extended for new data inputs and/or tasks, as different requirements arise. 

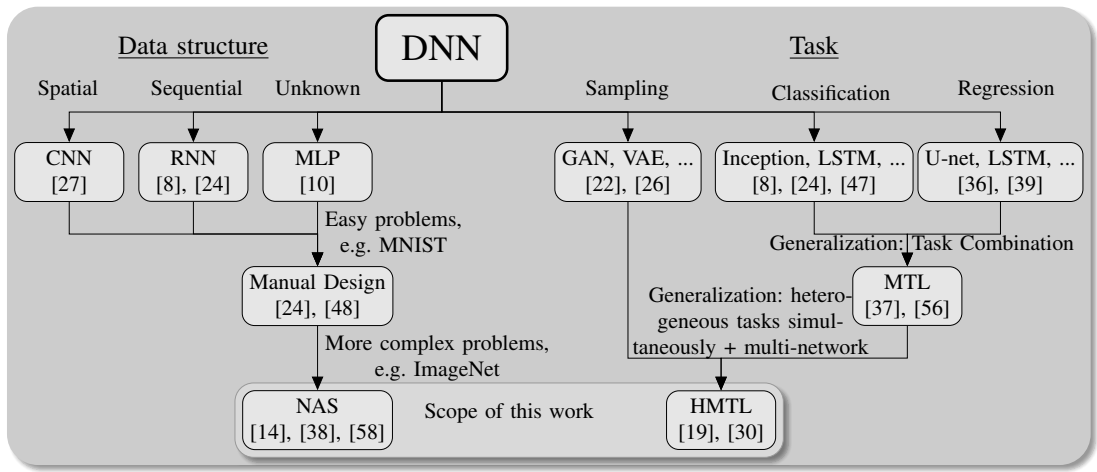
\begin{figure*}
	\begin{center}
		\resizebox{0.8\textwidth}{!}{
			\begin{tikzpicture}[every text node part/.style={align=center},arrow/.style = {thick,-stealth}]
			
			\newcounter{versep}
			\setcounter{versep}{2}
			
			\newcounter{minw}
			\setcounter{minw}{50}
			
			\newcounter{minh}
			\setcounter{minh}{25}
			
			\node at (-4, 0) (structure) {\large\underline{Data structure}};
			
			\node at (6, 0) (task) {\large\underline{Task}};
			
			\node at (0, 0)[rectangle, rounded corners=0.2cm,draw,minimum width=1.2*\value{minw},minimum height=1.2*\value{minh},fill=black!10, very thick] (dnn) {\LARGE DNN};

			\node at (-2, -\value{versep})[rectangle, rounded corners=0.2cm,draw,minimum width=\value{minw},minimum height=\value{minh},fill=black!10] (mlp) {MLP\\ \cite{cybenko_approximation_1989}};
			
			\node at (-4, -\value{versep})[rectangle, rounded corners=0.2cm,draw,minimum width=\value{minw},minimum height=\value{minh},fill=black!10] (rnn) {RNN\\ \cite{cho_learning_2014,hochreiter_long_1997}};
			
			\node at (-6, -\value{versep})[rectangle, rounded corners=0.2cm,draw,minimum width=\value{minw},minimum height=\value{minh},fill=black!10] (cnn) {CNN\\ \cite{krizhevsky_imagenet_2012}};
			
			\node at (3, -\value{versep})[rectangle, rounded corners=0.2cm,draw,minimum width=\value{minw},minimum height=\value{minh},fill=black!10] (generatives) {GAN, VAE, ...\\\cite{goodfellow_generative_2014,kingma_auto-encoding_2013}};
			
			\node at (6, -\value{versep})[rectangle, rounded corners=0.2cm,draw,minimum width=\value{minw},minimum height=\value{minh},fill=black!10] (classifyers) {Inception, LSTM, ...\\\cite{cho_learning_2014,hochreiter_long_1997,szegedy_inception-v4_2017}};
			
			\node at (9, -\value{versep})[rectangle, rounded corners=0.2cm,draw,minimum width=\value{minw},minimum height=\value{minh},fill=black!10] (regressers) {U-net, LSTM, ...\\\cite{mahendran_3d_2017,qin_dual-stage_2017}};

			\node at (-2, -2*\value{versep})[rectangle, rounded corners=0.2cm,draw,minimum width=\value{minw},minimum height=\value{minh},fill=black!10] (manual) {Manual Design\\\cite{hochreiter_long_1997,szegedy_going_2015}};
			
			\node at (7.5, -2*\value{versep})[rectangle, rounded corners=0.2cm,draw,minimum width=\value{minw},minimum height=\value{minh},fill=black!10] (mtl) {MTL\\\cite{meyerson_beyond_2017,zhang_regularization_2014}};

			\node at (-2, -3*\value{versep})[rectangle, rounded corners=0.2cm,draw,minimum width=\value{minw},minimum height=\value{minh},fill=black!10] (nas) {NAS\\\cite{elsken_neural_2019,miikkulainen_evolving_2019,zoph_neural_2016}};
			
			\node at (4.5, -3*\value{versep}) [rectangle,draw,minimum width=\value{minw},minimum height=\value{minh}, rounded corners=0.2cm,fill=black!10] (hmtl) {HMTL\\\cite{garciarena_towards_2021,li_person_2017}};
			
			\draw[->] (dnn.south) to[|-|] node[above right=2pt and -18pt] {Spatial} (cnn.north);
			\draw[->] (dnn.south) to[|-|] node[above right=2pt and -23pt] {Sequential} (rnn.north);
			\draw[->] (dnn.south) to[|-|] node[above right=4pt and -23pt] {Unknown} (mlp.north);
			
			\draw[->] (dnn.south) to[|-|] node[above right=2pt and -23pt] {Sampling} (generatives.north);
			\draw[->] (dnn.south) to[|-|] node[above right=2pt and -23pt] {Classification} (classifyers.north);
			\draw[->] (dnn.south) to[|-|] node[above right=2pt and -23pt] {Regression} (regressers.north);
			dashed
			\draw[->] (mlp.south) --  node[right= 0pt] {Easy problems,\\e.g. MNIST} (manual.north);
			\draw[->] (rnn.south) to[|-|] (manual.north);
			\draw[->] (cnn.south) to[|-|] (manual.north);
			\draw[->] (manual.south) -- node[right= 0pt] {More complex problems,\\e.g. ImageNet} (nas.north);
			
			\draw[->] (classifyers.south) to[|-|] node[below left=-3pt and -80pt] {Generalization: Task Combination} (mtl.north);
			\draw[->] (regressers.south) to[|-|] (mtl.north);
			
			\draw[->] (mtl.south) to[|-|] (hmtl.north);
			\draw[->] (generatives.south) -- + (0,-2) to[|-|]node[above right=-4pt and -44pt] {Generalization: hetero-\\geneous tasks simul-\\taneously + multi-network} (hmtl.north);
			
			\begin{scope}[on background layer]
			
			\node[fit=(hmtl)(nas),rectangle,fill=black!20,blur shadow={shadow blur steps=5}] (components){Scope of this work};
			
			\end{scope}
		
			\begin{scope}[on background layer]
				
				\node[fit=(cnn)(nas)(regressers)(dnn)(components),rectangle, rounded corners=0.5cm,fill=black!20,blur shadow={shadow blur steps=5}, minimum height=7cm, minimum width=11cm] (components){};

			\end{scope}
		\begin{scope}[on background layer]
			
			\node[fit=(hmtl)(nas),rectangle, rounded corners=0.3cm,fill=black!15,blur shadow={shadow blur steps=5},draw=black!40] (components){Scope of this work};
			
		\end{scope}
			\end{tikzpicture}}
		\caption{Evolution of DNN research. On the right-hand side of the figure, we can see how DNNs have been adapted to be able to satisfy different requirements, and how more complex fields have emerged by combining the different tasks. On the left-hand side, we can observe how the DNN structures have increasingly been specialized up to the point of requiring the application of NAS methods. This work lays at the top end of these two development branches.}
		\label{fig:diagram2}
	\end{center}
\end{figure*}

The diagram in Figure~\ref{fig:diagram2} shows a perspective on the directions in which the research of DNNs has evolved over the last few years according to two criteria: the structure type of the data they are specialized for, and the number and types of tasks they are designed for. 

Due to their incremented complexity, designing these models by hand poses an even harder challenge than the one of designing single-task models. Moreover, the lack of hand-made models expected to fulfill several assignments at the same time (unlike the case of single-task models) makes this problem even more difficult, as there are no references towards which the searches could be guided. Additionally, approaches based on reinforcement learning or gradient descent would require non-trivial adaptations in order to deal with the intricacies of operating in such a complex environment as that of HMTL. A clear example of these complexities is the necessity to satisfy data type constraints. Consequently, designing refined incremental NAS techniques that can cope with these restrictions becomes a necessity. More specifically, defining effective operators and smart approaches that maximize the efficiency of their application would be essential for obtaining powerful HMTL models.

This work aims at opening a new research line in the direction of making NAS more efficient. This has recently been identified as an area for potential improvement for these methods \cite{bender_understanding_2018,lu_nsga-net:_2018}, particularly for HMTL models. To that end, we attempt to illustrate the effectiveness of an \textit{intelligent} NAS approach by reducing the random component that characterizes some of the current structural search algorithms. With that goal in mind, we define a set of guidelines which can help a NAS algorithm to make an informed choice between all the variation operators at its disposal. These guidelines rely on a first step in which the model status is diagnosed using a set of metrics, which dictate the variation operator to be applied to improve the model in the second step. We attempt to make these guidelines as problem-independent as possible so that they ideally can be applied to any problem domain (e.g., classification, regression, generative modeling, or even combinations of them), regardless of the network architectures or variable dependencies involved. In that regard, we identify the areas in which the modeling difficulty is high: problems that fit the HMTL paradigm \cite{garciarena_towards_2021,li_person_2017}. To illustrate the effectiveness of these guidelines in complex domains, we evaluate them on NAS algorithms for models that satisfy two particular characteristics: 1) They comprise multiple sub-networks that interact with each other. 2) They solve multiple heterogeneous machine learning problems (e.g., classification, regression, and data generation simultaneously). More specifically, we employ the VALP \cite{garciarena_towards_2021} (a recently introduced HMTL model) as a testbed to demonstrate the effectiveness of the proposed strategy.

This paper is organized as follows. In the next section, literature relevant to this work is covered. The problem in which the approach is tested is introduced in Section~\ref{sec:valp}. In Section~\ref{sec:approach}, the ideas which are the main contribution of this work are described. These ideas are then materialized into mechanisms for improving NAS runs, which are described in Section~\ref{sec:search}. The experiments designed for showing the potential of the proposal are presented in Section~\ref{sec:experiments}, and the obtained results are summarized and discussed in Section~\ref{sec:results}. Finally, Section~\ref{sec:conclusions} contains the conclusions drawn from this work, as well as some future research lines.


\section{Related Work}

As stated in the introduction, the approach presented in this work consists of the smart application of variation operators in order to increase the efficiency level of NAS algorithms, particularly -but not limited to- when applied to HMTL. In this section, we first introduce relevant work on the NAS area, classified by the type of search and operators they employ, so that the beneficial aspects of each type are identified. By integrating multiple perspectives on the NAS problem, an algorithm with access to multiple options from which to choose when making decisions could be designed. We next discuss some work performed on DNN model diagnosis, as the metrics proposed in these works will be useful to make an estimation of the relevance of each component within a neural model. This estimation can ultimately lead to informed decisions among all the possibilities imported from the different NAS approaches. Finally, we cover some different approaches to the HMTL problem, a problem definition to which the efforts in this paper are devoted, and which is more general than the commonly addressed single-task scenarios. By proving the proposal of this paper in such a complex environment as HMTL, it seems fair to assume that it could also be successfully applied to other \textit{simpler} scenarios - such as single task ones.

\subsection{Neural architecture search}

As previously mentioned in the introduction, several techniques have influenced multiple NAS methods. We limit this review to the approaches that have the largest influence on the proposed set of guidelines: neuroevolution (EAs which usually rely on neural variation or mutation operators) and network morphism (NM).

\subsubsection{Neuroevolution}

In the last couple of decades, several works have used evolutionary approaches for NAS with the ultimate goal of producing DNNs able to cope with different tasks. Although these works have particularly focused on image-related tasks, many of the proposals could be applied to other types of tasks.

The traditional approach to neuroevolution (NE) commonly considers relatively low-parametrized networks, both regarding the number of layers and the number of neurons in them. However, as the hardware supporting DNNs has improved, these methods have shifted from performing low-level modifications, e.g., the addition of one neuron or connection \cite{stanley_compositional_2007,stanley_evolving_2002}, to more complex operations, like the concatenation of full neural cells to the DNN \cite{liang_evolutionary_2018,lu_nsga-net:_2018}. This second kind of evolution has proven competitive against hand-crafted structures and is the most popular approach considering the amount of recent work devoted to it. Currently, these two scopes of variation operators are known as micro (modifications limited to the small cells or sub-networks within the model) and macro search (altering the general structure of the neural model) \cite{white_local_2020}.

The work in \cite{real_regularized_2019} presents an NE approach that adopts the NASNet search space (initially designed for a reinforcement learning-based NAS) \cite{zoph_learning_2018}. The authors propose the incorporation of an age property for all individuals in an NE procedure carried out in the NASNet space, in order to favor individuals of recent creation at the time of performing the tournament selection. 

Some other approaches, while still being framed in the image treatment scheme, have variations especially relevant to our work. For example, in \cite{lu_nsga-net:_2018}, NSGA-Net is proposed. This population-based algorithm permits the inclusion of (potentially conflicting) objectives as opposed to the classic single error metric-minimization scenario. This way, the authors address the problem of low efficiency on state-of-the-art NE algorithms by introducing a second objective which seeks the minimization of the computational complexity of the models. 
The work presented in \cite{chen_modulenet_2020} introduces ModuleNet. This NE algorithm is largely inspired by \cite{lu_nsga-net:_2018}, although new mutation operators are introduced. This NE algorithm is based on connecting sub-networks found in top-performing DNNs proposed in the literature. 

The previously reviewed works (as well as many others not included here) mostly follow the same pattern: They introduce a framework different from those that have already been proposed, and design ad-hoc operators for it. In this work, we aim at forming a set of guidelines which are able to operate in different schemes in terms of model structure and problem domain by encompassing different types of operators and applying them when their positive impact towards the search can be maximized.

\subsubsection{Network Morphism}

Another research subfield which has grown separately, but is related to NE due to being based on structural variation operators, is network morphism. It consists of a special set of operations for extending DNN structures in such a way that the performance of the network is not altered.

The work described in \cite{chen_net2net:_2015} proposes two operators for expanding DNN architectures, Net2Net, applicable to both MLP and CNN architectures. Their effectiveness is tested in a framework in which, initially, a relatively shallow and narrow \textit{teacher} network is trained for the objective task. Next, one of the two operators; \texttt{Net2WiderNet} (enlarging the size of a layer) or \texttt{Net2DeeperNet} (introducing a new layer to the DNN), are applied to increment the number of parameters of the model. By conveniently initializing the added weights and/or the activation functions, the newly created \textit{student} network is able to produce the same result as the \textit{teacher} network. However, because its modeling power has increased, better results can be expected with further training.

These operators are first employed to gradually transform a shallow \textit{teacher} into an inception network. Results show higher accuracy and faster convergence than training the same structure from scratch. 

The authors of \cite{wei_network_2016} then extended that work by resolving some of its inherent limitations: e.g., the inclusion of non-idempotent activation functions in the network modifications. 
Besides, the subnet adding operator was also presented, which is equivalent to adding several layers at once. The authors finally used the \textit{network morphism} term to name the framework containing these kinds of operators. 
This extension of \textit{Net2Net} is able to outperform the original proposal both in learning speed and final accuracy.

The work in \cite{elsken_simple_2017} takes full advantage of the framework defined in \cite{wei_network_2016} and uses it as a tool for a Neural Architecture Search by Hillclimbing (NASH), using a simple structure as the starting point. 

To the authors' knowledge, the research carried out using these operators is rather limited, considering the complementary role they can play for the more common operators usually employed for NAS. Therefore, we integrate them into the NAS framework governed by the guidelines proposed in this work, along with more traditional operators for NAS algorithms.

\subsection{Model internal diagnosis}

Studies attempting to understand the way DNNs operate have yielded many interesting approaches \cite{alain_understanding_2016,arras_explaining_2017,shwartz-ziv_opening_2017}.

The authors of \cite{shwartz-ziv_opening_2017} propose the diagnosis of a neural model by computing the mutual information between the representations of the information found in the different layers of a DNN, and the input and output of the network during the training of the model. They concluded that, when trained with the common combination of stochastic gradient descent (SGD) and backpropagation, the weight optimization procedure of a DNN consists of two different phases, the information compression phase and the error minimization period.


In \cite{alain_understanding_2016}, a similar approach is presented. The information representation in each layer of a classification DNN is extracted for a set of observations, and a linear classifier is fitted between each of these representations and the original classes, independently. The errors reported by the classifiers from the different layers can serve as a measure of the quality of the information representation at each level of depth in a DNN. Because linear classifiers are rather limited and require rich representations of the data to perform well, it can be expected that, the deeper the layer -and therefore, the richer the representation-, the better a linear classifier will perform.

In the both previous approaches, \cite{alain_understanding_2016,shwartz-ziv_opening_2017}, comparing the values given by the metrics (the mutual information and the classification error) between layers can help to understand the level of importance of each layer within the general model context.

In an attempt to identify the origin of an issue in the DARTS \cite{liu_darts_2018} search space, the authors of \cite{wang_rethinking_2021} propose the deletion of different parts of a DNN and using the observed performance decrease of the overall DNN as an estimation of the relevance of that part to the overall model.

Although the explainability of the decisions made by DNN models is not among the objectives of this work, trying to estimate the relevance of a sub-network (named sub-DNN indistinctly in this work) within a neural model composed of multiple sub-networks is. In this work, we propose the exploitation of this kind of metrics so as to assess the level of importance of each sub-DNN in the general model in order to determine the structural variations which have larger potential for improvement in terms of model performance.

\subsection{Heterogeneous learning}

In \cite{garciarena_towards_2021}, a framework for HMTL was defined, called VALP, which will be introduced in more detail in Section~\ref{sec:valp}. 
One particular characteristic of the VALP is the fact that the structure of the graph is not fixed, and it is therefore optimizable. That work makes an initial exploration of the search space of VALPs by employing a random search and comparing its results to a fixed regular MTL model. That experiment showed to which extent finding the right structure is an important aspect for the performance of the HMTL models. This observation later motivated an extension of this work in \cite{garciarena_automatic_2020}, where a VALP structural search algorithm was proposed. The method consists of four different operators which modify the structure of a VALP used as a base for a Hill Climbing (HC) algorithm. All four operators are based on connection modification, namely, add and delete connection, and insert and delete network. Conducted experiments showed that, even though the operator to be applied was randomly chosen, the HC was able to find better performing structures than a simple random search.  

After discovering the potential of a search algorithm on an HMTL environment, in this paper, we go one step further in this area by exploiting NAS methods in an HMTL framework. Our main goal is, as we recognize the difficulty of dealing with multiple inputs, outputs, loss functions, sub-DNN types, etc., to present advances to the NAS field that can make the structural search feasible even in such convoluted search spaces by presenting a set of guidelines that can be used by NAS algorithms. Although its initial presentation is directed towards the HMTL framework, the application of these guidelines is not bounded to that environment, as it can be exported to many other NAS formulations.



Multi-loss function joint learning (JL) \cite{li_person_2017} was initially defined over the \textit{person re-identification problem}, which consists of matching identity classes in detected person bounding boxes from non-overlapping camera views. To that end, several different features are learned for each person in multiple images, and are afterwards used to match persons across images according to the similarity of these features. The authors choose to employ a JL model to test the hypothesis of whether learning these features can yield better results if done together rather than using separate models for each one. With that goal in mind, the authors manually build a CNN in which the first part is shared, before the network is divided into two branches, each of which has a specific design for the different tasks and is trained using a separate loss function. 

DG-Net \cite{zheng_joint_2019} introduces a new dimension to the JL framework as it was designed to fulfill tasks different from each other. In contrast to \cite{li_person_2017}, DG-Net is able to both classify and generate new samples similar to those in the training set. DG-Net is designed as a module-based neural model, in which each module has a specific task to fulfill, varying from encoding or decoding image features or structures, to discriminating the images based on the mentioned features. Because each part of the model has a predefined role, the structure of the model cannot be changed, as it could mean a change on the role played by each sub-network of the model. The training of the model for performing such diverse tasks requires, of course, the management of different loss functions. The components and loss functions are arranged in a predefined structure to maximize the modeling capabilities of each of the sub-networks within DG-Net. In order to combine the losses in a single expression, a weighted average of them is computed by hand-picking weights.

\section{The heterogeneous, module-based model}\label{sec:valp}

We aim at defining a set of guidelines as general as possible capable of guiding a NAS algorithm and consequently choose an application benchmark which shares that characteristic: a model which is able to handle multiple data inputs, can make predictions for more than one output at time (be they homogeneous or heterogeneous), and is composed of sub-DNNs itself. The inputs, outputs, and modules within the model are interconnected with a scheme of connections which points which information (either from a model input or a sub-network) is redirected to which place (either another sub-network or a model output), forming a \textit{directed graph}. Each output corresponds with a prediction required to the model. Because the performance of a model can be assessed by (at least) as many metrics as outputs it has, this problem has a multi-objective nature.

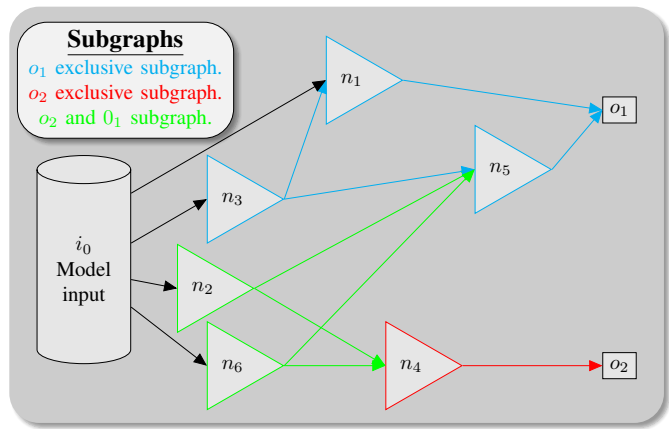
\begin{figure}
	\begin{center}
		\resizebox {8.8cm} {!}{
			\begin{tikzpicture}
			
			\node at (3,2.3) (datatag) [cylinder, draw,shape border rotate=90,aspect=0.3, minimum height=3.5cm,fill=black!10] {\begin{tabular}{c}$i_0$\\Model\\input\\\end{tabular}};
			
			\node at (3.7,5.5) (legend) [rectangle, draw,rounded corners=0.5cm,blur shadow={shadow blur steps=5}, fill=black!5] {\setlength{\tabcolsep}{2pt}\begin{tabular}{c}
				\textbf{\underline{\large Subgraphs}}\\
				\textcolor{cyan}{$o_1$ exclusive subgraph.}\\
				\textcolor{red}{$o_2$ exclusive subgraph.}\\
				\textcolor{green}{$o_2$ and $0_1$ subgraph.}\\
				\end{tabular}};

			\node at (5.5,0.7)[regular polygon,regular polygon sides=3,draw=green,shape border rotate=-90,fill=black!10] (n8) {$n_6$};

			\node at (7.5,5.5)[regular polygon,regular polygon sides=3,draw=cyan,shape border rotate=-90,fill=black!10] (n1) {$n_1$};
			
			\node at (8.5,0.7)[regular polygon,regular polygon sides=3,draw=red,shape border rotate=-90,fill=black!10] (n7) {$n_4$};
			\node at (5,2)[regular polygon,regular polygon sides=3,draw=green,shape border rotate=-90,fill=black!10] (n2) {$n_2$};
			\node at (5.5,3.5)[regular polygon,regular polygon sides=3,draw=cyan,shape border rotate=-90,fill=black!10] (n3) {$n_3$};
			
			\node at (10,4)[regular polygon,regular polygon sides=3,draw=cyan,shape border rotate=-90,fill=black!10] (n5) {$n_5$};
			
			\node at (12,5)[rectangle, draw,fill=black!10] (o1) {$o_1$};
			\node at (12,0.7)[rectangle, draw,fill=black!10] (o2) {$o_2$};

			\draw[->,green] (n2.east)--(n7.west);
			\draw[->,green] (n2.east)--(n5.west);
			
			\draw[->,cyan] (n3.east)--(n1.west);
			\draw[->,cyan] (n3.east)--(n5.west);
			
			\draw[->] (datatag)--(n8.west);
			\draw[->] (datatag)--(n3.west);
			\draw[->] (datatag)--(n2.west);
			\draw[->] ([yshift=1.3cm]datatag.east)--(n1.west);
			\draw[->,green] (n8.east)--(n5.west);
			\draw[->,green] (n8.east)--(n7.west);
			
			\draw[->,red] (n7.east)--(o2.west);
			
			\draw[->,cyan] (n5.east)--(o1.west);
			
			\draw[->,cyan] (n1.east)--(o1.west);
			
			\begin{scope}[on background layer]
			
			\node[fit=(o1)(o2)(datatag)(n1),rectangle,fill=black!20, rounded corners=0.5cm,blur shadow={shadow blur steps=5}, minimum height=7cm, minimum width=11cm] (components){};
			
			\end{scope}
			
			\end{tikzpicture}}
		\caption{Example of a VALP, with its different subgraphs. The $n_1$, $n_3$, and $n_5$ sub-DNNs are part of the exclusive subset of $o_1$, as they are exclusively connected to $o_1$. Similarly for the $n_4$ sub-network and $o_2$. Adding the $n_2$ and $n_6$ sub-DNNs to either subset would result in $o_1$ and $o_2$ subsets respectively, as these would contain all the sub-networks involved in these outputs, including sub-DNNs which are involved in other outputs.}
		\label{fig:subpath}
	\end{center}
\end{figure}

The recently proposed VALP \cite{garciarena_towards_2021} is an example of a module based model used for solving HMTL problems. Its structure can be defined as a directed graph, $G=(V,A)$ in which the sub-DNNs within the model are represented by the vertices $V$, and the flow of information produced and received by them is directed by the arcs $A$. We name the combination of these two sets as the set of components. The vertices in $V$ can be categorized in three different disjoint subsets; $I \cup N \cup O = V$. $I$ is composed of the source nodes $i_j$, and, in the network topology, these are the sources of information, the data. $N$ contains the internal nodes $n_k$, which are the actual sub-networks within the VALP, each of which can be defined by different architectures and hyperparameters. Finally, $O$ contains the set of sink nodes $o_l$, the final components in which the final predictions (regarding the data present in the source nodes) of the VALP can be collected. An example of a VALP can be seen in Figure~\ref{fig:subpath}.

The VALP can be trained using backpropagation, a gradient descent algorithm, and a set $L$ of loss functions (which, in this work, are combined by addition), containing at least one loss function for each item in $O$.

In a VALP (as in any other module-based model with multiple inputs, outputs, and components), the sub-networks can be grouped according to different criteria. We define the following subgraphs of $G$: \textit{output subgraph}, and \textit{output exclusive subgraph}. The \textit{subgraph} of an output $o_l$ consists of all the components that, upon modification, alter the prediction in $o_l$. The \textit{output exclusive subgraph} consists of a similar subgraph, although in this case, the components that affect multiple outputs are not included in either \textit{output subgraph}.

Figure~\ref{fig:subpath} shows two examples of subgraphs. The first one, colored in blue, is the exclusive subgraph correspondent to $o_1$. It contains three networks ($n_1$, $n_3$, $n_5$), all of them only ultimately connected to $o_1$. $n_2$, and $n_6$ cannot be part of that exclusive subgraph because they also provide data to $o_2$. The exclusive subgraph correspondent to $o_2$, in red, is composed of just $n_4$, as it is the only network exclusively connected to that output. The output subgraph of $o_1$ is composed of all the components ultimately connected to $o_1$, i.e., $n_1$, $n_2$, $n_3$, $n_5$, and $n_6$. Finally, $n_4$, $n_6$, and $n_2$ would form the output subgraph of $o_2$.


The first thing to be defined in a VALP are the $I$ and $O$ sets. Next, $A$ and $N$ are defined to design the architecture of the model. In this work, this is achieved by employing structural optimization methods. Depending on their implementation, some optimization algorithms (e.g., those based on variation operators) could try to evaluate solutions which fall outside the constraints of the problem, which would be a poor exploitation of computational resources. Although techniques for detecting and \textit{repairing} this kind of solutions exist \cite{coello_theoretical_2002}, a common approach when facing such complex search spaces is to rely on operators which guarantee that their product is going to be feasible. In this work, we choose the second option. We understand a structurally correct VALP as one which guarantees to comply with a set of characteristics, which necessarily include:

\begin{itemize}
	\item The model has as many model outputs as the problem has targets.
	\item Correspondence in terms of data type between each target variable and each model output has to exist. For example, a classification and a regression output do not share the same characteristics (the neurons in a classification output have to add to one, and are therefore activated by the softmax function).
	\item Every sub-DNN and model output must receive data from at least another component, and the data coming from every sub-network and model input must go to at least another component.
	\item This instance of VALP model does not consider the possibility of recurrent connections, despite this limiting the capacity of the model to deal with temporal data. However, the proposal of this work would still be valid if recurrent connections were contemplated.
\end{itemize}

\section{Intelligent search}\label{sec:approach}


Due to the costly nature of some NAS algorithms caused by the magnitude of the search space, an efficient structural search of module-based models is crucial, especially when dealing with HMTL problems. This efficiency is mainly dependent on the operators integrated on the search algorithm, and, most importantly, how they are employed. In this section, critical aspects of different search methods are identified, before reflecting on how to exploit these characteristics in order to improve the efficiency of the search algorithms. 

First, we categorize the search methods by the number of neural models being taken into account in any given moment.

\begin{itemize}
	\item Single model search: In this instance, the search algorithm consists of improving a single model at a time, as in a local search (e.g., hill climbing).
	\item Population based search: This second formulation considers several models at each time during the search.
\end{itemize}

\subsection{Model internal diagnosis}

The first key question in the proposed intelligent structural search is identifying \textit{which} component or part of the structure (in our case, a sub-network) should be improved at a given point. To that end, we identify diverse sources of information depending on the type of search, which could help making the right decision in this matter.

In a single model scenario, the main sources of information consist of:

\begin{enumerate}
	\item Comparisons with the performance of models evaluated in previous iterations of the algorithm.
	\item The relevance of the different components within the model to the final predictions made by the model.
	\item The effectiveness of the training procedure to improve the prediction.
\end{enumerate}

In the population-based search, along with these three information sources, other information sources are also available. 
These can be used to gauge the performance of a given model with respect to its peers, which can provide a more accurate idea of which component of the model, when modified, can provide a better gain in terms of model performance or loss function optimization.

\subsection{Metrics} \label{sec:metrics}

With the information sources identified, the second step is to determine how the knowledge is going to be captured. Focusing on the single model scenario, we formalize four different metrics:

\begin{enumerate}
	\item \emph{Historic sub-loss information:} Performance metrics extracted from the loss functions associated to the sub-network. The performance of a sub-DNN (o set of them) along time can be estimated comparing and combining metrics -e.g., the loss function values- at each iteration.
	
	\item \emph{Module intervention:} Inspired by the techniques of intervention for causal discovery \cite{eberhardt_interventions_2007} and neural architecture selection methods \cite{wang_rethinking_2021}, here we modify some component of a module (e.g., setting the weights to random values) to estimate its importance as the gain/loss in performance as a result of the intervention.
	
	\item \emph{Input intervention:} Similarly, it is possible to intentionally modify input values (i.e., a subset of the features of the data) and estimate their relevance with respect to the predictions.
	
	\item \emph{Dependency measures:} Metrics (such as the mutual information or a classification algorithm error \cite{alain_understanding_2016}) between the output of each sub-network and the model output(s) it contributes to, would ideally improve compared to the output of the sub-DNN preceding it \cite{shwartz-ziv_opening_2017}. If this is not the case, it could be interpreted that the component is not helpful.
	
\end{enumerate}

For population-based approaches, we define an additional metric based on comparisons between models in population-based searches (although it could also be applied to the isolated model search by comparing the current model with other models in previous stages of the search).

\begin{enumerate}
	\setcounter{enumi}{4}
	\item \emph{Relative performance:} several rankings -at least one per output- can be arranged, according to the performance of the model in each output, relative to the rest of models. The position of a model in the ranking of a given output determines the quality of the subgraph of that output.
	
\end{enumerate}


\subsection{Variation operator types}\label{sec:mutators}

The third step is to define variation operators that cover the different needs that the models can present at different points during their development.

In this work, we categorize the variation operators according to two attributes: their \textit{aggressiveness}, and the effect they have on the complexity of the model. Regarding aggressiveness, we distinguish two types of mutators:

\begin{enumerate}
	\item We consider an operator to be \textit{aggressive} when it performs drastic alterations to the model structure, in such a way that the performance of the model can be severely changed in at least one of the objectives (e.g., operators commonly used in NAS).
	\item On the contrary, an operator is considered as \textit{gentle} in case the performance of the model does not vary after its application (i.e., morphism operators).
\end{enumerate}

When discriminating operators by their effect on the complexity of the model, we also divide the set of mutators into two subsets;

\begin{enumerate}
	\item An operator is considered a \textit{reducer} when its application decreases the number of weights in the model, and thus, theoretically, the modeling capacity.
	\item Alternatively, an operator is an \textit{extender} when the model sees its number of weights increased.
\end{enumerate}

Because we have two categories for each characteristic, we can define four type combinations. First, an aggressive extender operator would increase the number of weights of a model at the same time as the performance of the model is altered. For example, integrating a new random sub-network to a subgraph of an output could alter the performance of the model in that output. 

Secondly, a gentle extender operator would increase the modeling capacity of the model without modifying the performance of the model, e.g., by modifying other components already present in the model and cautiously designing and placing the new component.

Thirdly, an aggressive reducer would decrease the modeling capacity and have the collateral effect of altering the model performance, e.g., by deleting a sub-DNN or a connection which was relevant to the overall model.

Finally, a gentle reducer would delete certain parts of a model, without affecting the performance of the model. The deleted parts would need to be irrelevant to the model.

\subsection{Donation operator}\label{sec:donation}

In population-based searches, mutation operators are not the only method to perform alterations to models. In this case, although they have been widely omitted by the NE community \cite{galvan_neuroevolution_2020,uriot_safe_2020}, we define a special version of crossover, traditionally referenced as the \textit{conjugation} operator \cite{harvey_microbial_2009}. In this method, a \textit{donor} model donates a part of itself (e.g., the output exclusive subgraph) to a \textit{host} model. 

\subsection{Principles for using the metric information}

In this section, we propose a set of criteria aiming at optimizing NAS procedures for HMTL models (although their application is not limited to that kind of models), exploiting the metrics defined in Section~\ref{sec:metrics} to guide the selection of the variation and donation operators, as defined in Section~\ref{sec:mutators} and Section~\ref{sec:donation}.

\subsubsection{Historic sub-loss information}

This metric can be used to observe the behavior of one or more model outputs by fitting a linear regression model which attempts to predict the sub-loss value of an output, given the training step. This way, the slope of the loss function can be approximated with a line and, depending on that value, different approaches can be taken:

\begin{itemize}
	\item If the slope is close to 0 or positive, it can be concluded that the output has converged. In that case, an aggressive operator could take the model away from that local optima.
	\item When the slope is slightly smaller than 0, it can be interpreted that the output is still improving, although a major improvement is unlikely. In this case, a gentle extender operator could add more modeling power, helping the model perform another significant gain without losing the current performance.
	\item In the case in which the slope is considerably smaller than 0, the output is still in the early phase of improvement, and should be left as it is until a certain level of convergence is reached, i.e., the previous two scenarios.
\end{itemize}

\subsubsection{Module intervention}

This metric can be used to determine the relevance of a given sub-network to the overall model by measuring the performance loss after resetting the weights of that sub-DNN.

\begin{itemize}
	\item If the performance loss is not great for any output, the importance of the sub-DNN to the model is low, and a reducing operator could be advised.
	\item On the contrary, if the drop off is significant, the component is assumed to be relevant, and should either remain intact or be expanded using a gentle operator.
	\item Finally, if the sub-network is connected to an output which was not affected, a connection deletion would reduce the model complexity without deteriorating the overall model performance.
\end{itemize}

\subsubsection{Input intervention}

Similarly to Module intervention, this metric would estimate the importance of a given input to the final prediction of the model. This could be done by observing the performance change in the different outputs when randomly changing a subset of the features of the data:

\begin{itemize}
	\item If the performance loss is not great, then the input is not very relevant to the output, and deleting a connection that connects the path between the input and the prediction would be advisable, so that the model graphically represents that \textit{independence}.
	\item If a significant percentage of performance is lost, then the input is relevant to the output, and no connection should be deleted.
\end{itemize}

\subsubsection{Dependency measures}

As was the case for the module intervention, this metric serves the purpose of measuring the importance of a component for the model. In this case, a metric (e.g., the mutual information or the error of a linear estimator) is computed between a model output and the outputs of the components on its subgraph. Next, for each sub-network, the obtained value is compared to the values of its predecessor in the model.

\begin{itemize}
	\item When the measure indicates a larger dependency between the values, it can be assumed that the component is performing satisfactorily, and should be either gently expanded or left unchanged.
	\item If the value does not improve, the component is not performing as expected, and a reducer operator can be applied without losing much potential.
\end{itemize}

\subsubsection{Relative performance}

By constructing rankings of models according to their performance in the different outputs, it would be possible to estimate the relative performance of a model in that output. A model with all but one output in the higher part of their corresponding rankings could become the host of the exclusive subgraph of a model with a high rank in that specific output ranking. This vision is closely related to multi-objective optimization, as one model can be viewed as valuable or useless depending on different factors, as the output being evaluated, or the current state of the search.

\section{Searching for optimal VALP structures using variation operators}\label{sec:search}

The previous section presented a general approach and guidelines towards an intelligent structural search. In order to show its utility, this theoretical framework is implemented into the VALP NAS context. In what follows, we introduce variation operators which can be applied to the VALP, but could, generally, be applied to any other neural model based on sub-DNNs (or neural cells). We decided that the defined operators must comply with the characteristic of having to produce structurally valid VALPs. The operators are classified according to the characteristics described in Section~\ref{sec:classicMut} (aggressiveness and effect over the complexity of the model) and the scopes of application:

\begin{enumerate}
	\item Sub-networks, operators used in micro searches.
	\item General model structure, macro search operators used for modifying the connections between sub-DNNs.
	\item Hyperparameters
	\item Crossover operator
\end{enumerate}


\subsection{Sub-networks}

We start with the operators with the most reduced performance scope (micro search): layer-wise modifications of the sub-networks in a VALP. Three different mutation operators have this scope:

\begin{itemize}
	\item \texttt{add\_layer}: This extender operator adds a layer in the network. Depending on how the weights are initialized and where the layer is added, this operator can be aggressive (e.g., by randomly initializing the weights) or gentle (e.g., by using the morphism approach).
	\item \texttt{remove\_layer}: This operator deletes a layer from the network. The rest of the layers remain the same. As a reducer, this operator is aggressive.
	\item \texttt{extend\_layer}: This operator adds neurons to a layer from the network. The remaining layers stay the same. This extender operator can be aggressive or gentle.
\end{itemize}

\subsection{General model structure} \label{sec:classicMut}

The next set of operators is capable of affecting the VALP structure in its higher level (macro search), i.e., the interconnections between the different sub-DNNs in a VALP. We define five modifiers with this capacity: 

\begin{itemize}
	\item\texttt{add\_connection}: Given two currently unlinked sub-networks of a VALP, this operator links them by creating a new connection. In other words, the second sub-DNN receives the output of the first sub-network as additional input. This extender operator can be both gentle or aggressive.
	\item\texttt{delete\_connection}: Given a connection of a VALP, this operator deletes it. This operator is aggressive and reducer.
	\item\texttt{insert\_network}: Given a connection of a VALP, this operator inserts a network in the middle of the connection. For example, if a connection $c_0$ that links $n_0$ to $n_1$ is chosen, a connection $c_1$ between $n_0$ and the newly created $n_m$, and a connection $c_2$ between $n_m$ and $n_1$ are created, and $c_0$ is deleted. This expander operator can be both aggressive or gentle.
	\item\texttt{delete\_network}: Given a network $n_m$ of a VALP, this operator deletes it. Each sub-network providing data to $n_m$ switches to supplying data to each and every sub-DNN $n_m$ provided data to. This operator is reducer, and, as determined by an additional experiment reported in the supplementary material of this paper, aggressive.
	\item \texttt{clone\_network}: Given a network of a VALP, this operator duplicates that network and all the connections related to it. This operator is an expander and can be both gentle or aggressive. Applying a 0.5 factor to the outputs of both the original and the clone networks neutralizes the effect of the operator on the outputs, resulting in a gentle operator.
\end{itemize}

These last five methods will be applied only if structural correctness is guaranteed. For example, \texttt{delete\_connection} will not, under any circumstances, delete a connection when it is the only source of data of a sub-network, or \texttt{delete\_network} will never suppress a sub-network when it is the only one between a model input and a model output. 

The gentle operators defined in this work depend entirely on reusing and adequately modifying the weights optimized in the previous training epochs. We reuse the weights learned by a model before being altered, i.e., we apply weight inheritance, whenever it is viable (when a random sub-DNN is added to a VALP, no weight inheritance is possible).

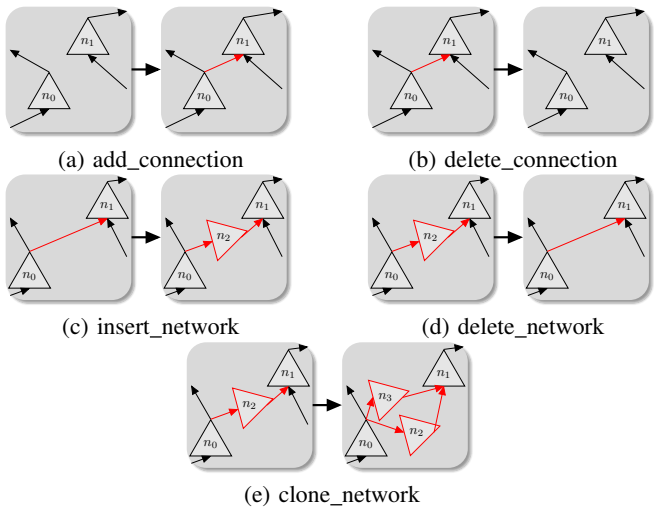
\begin{figure}
	\begin{center}
		\begin{subfigure}[b]{0.22\textwidth}
			\resizebox {3.8cm} {!}{
				
				\begin{tikzpicture}
				
				\coordinate (i0) at (-1,-1);
				\coordinate (o0) at (-1,1);
				\coordinate (i1) at (2, 0);
				\coordinate (o1) at (2,2);
				
				\node at (0,-0.2)[regular polygon,regular polygon sides=3,draw,fill=black!10,inner sep=1pt] (n0) {$n_0$};
				\node at (1,1.2)[regular polygon,regular polygon sides=3,draw,fill=black!10,inner sep=1pt] (n1) {$n_{1}$};
				
				\draw[->] (i0.north)-- (n0.south){};
				\draw[->] (n0.north)-- (o0.south){};
				\draw[->] (i1.north)-- (n1.south){};
				\draw[->] (n1.north)-- (o1.south){};

				\coordinate (i01) at (3,-1);
				\coordinate (o01) at (3,1);
				\coordinate (i11) at (6, 0);
				\coordinate (o11) at (6,2);
				
				\node at (4,-0.2)[regular polygon,regular polygon sides=3,draw,fill=black!10,inner sep=1pt] (n01) {$n_0$};
				\node at (5,1.2)[regular polygon,regular polygon sides=3,draw,fill=black!10,inner sep=1pt] (n11) {$n_{1}$};
				
				\draw[->] (i01.north)-- (n01.south){};
				\draw[->] (n01.north)-- (o01.south){};
				\draw[->] (i11.north)-- (n11.south){};
				\draw[->] (n11.north)-- (o11.south){};
				\draw[->] (n01.north)-- (n11.south)[red]{};
				
				
				\begin{scope}[on background layer]
				
				\node[fit= (i0)(o1),rectangle,fill=black!15, rounded corners=0.5cm,blur shadow={shadow blur steps=5}] (pre){};

				\end{scope}
				
				\begin{scope}[on background layer]
				
				\node[fit= (i01)(o11),rectangle,fill=black!15, rounded corners=0.5cm,blur shadow={shadow blur steps=5}] (post){};

				\end{scope}
				
				\draw[->] (pre)-- (post)[ultra thick]{};
				
				\end{tikzpicture}}
			\caption{add\_connection}
			\label{fig:add_conn}
		\end{subfigure}
		\qquad
		\begin{subfigure}[b]{0.22\textwidth}
			\resizebox {3.8cm} {!}{
				\begin{tikzpicture}
				
				\coordinate (i0) at (-1,-1);
				\coordinate (o0) at (-1,1);
				\coordinate (i1) at (2, 0);
				\coordinate (o1) at (2,2);
				
				\node at (0,-0.2)[regular polygon,regular polygon sides=3,draw,fill=black!10,inner sep=1pt] (n0) {$n_0$};
				\node at (1,1.2)[regular polygon,regular polygon sides=3,draw,fill=black!10,inner sep=1pt] (n1) {$n_{1}$};
				
				\draw[->] (i0.north)-- (n0.south){};
				\draw[->] (n0.north)-- (o0.south){};
				\draw[->] (i1.north)-- (n1.south){};
				\draw[->] (n1.north)-- (o1.south){};
				\draw[->] (n0.north)-- (n1.south)[red]{};
				
				\coordinate (i01) at (3,-1);
				\coordinate (o01) at (3,1);
				\coordinate (i11) at (6, 0);
				\coordinate (o11) at (6,2);
				
				\node at (4,-0.2)[regular polygon,regular polygon sides=3,draw,fill=black!10,inner sep=1pt] (n01) {$n_0$};
				\node at (5,1.2)[regular polygon,regular polygon sides=3,draw,fill=black!10,inner sep=1pt] (n11) {$n_{1}$};
				
				\draw[->] (i01.north)-- (n01.south){};
				\draw[->] (n01.north)-- (o01.south){};
				\draw[->] (i11.north)-- (n11.south){};
				\draw[->] (n11.north)-- (o11.south){};
				
				
				\begin{scope}[on background layer]
				
				\node[fit= (i0)(o1),rectangle,fill=black!15, rounded corners=0.5cm,blur shadow={shadow blur steps=5}] (pre){};

				\end{scope}
				
				\begin{scope}[on background layer]
				
				\node[fit= (i01)(o11),rectangle,fill=black!15, rounded corners=0.5cm,blur shadow={shadow blur steps=5}] (post){};

				\end{scope}
				
				\draw[->] (pre)-- (post)[ultra thick]{};
				
				\end{tikzpicture}}
			\caption{delete\_connection}
			\label{fig:delete_conn}
		\end{subfigure}
		\qquad
		\begin{subfigure}[b]{0.22\textwidth}
			\resizebox {3.8cm} {!}{
				\begin{tikzpicture}
				
				\coordinate (i0) at (-1,-1);
				\coordinate (o0) at (-1,1);
				\coordinate (i1) at (2, 0);
				\coordinate (o1) at (2,2);
				
				\node at (-0.5,-0.5)[regular polygon,regular polygon sides=3,draw,fill=black!10,inner sep=1pt] (n0) {$n_0$};
				\node at (1.5,1.3)[regular polygon,regular polygon sides=3,draw,fill=black!10,inner sep=1pt] (n1) {$n_{1}$};
				
				\draw[->] (i0.north)-- (n0.south){};
				\draw[->] (n0.north)-- (o0.south){};
				\draw[->] (i1.north)-- (n1.south){};
				\draw[->] (n1.north)-- (o1.south){};
				\draw[->] (n0.north)-- (n1.south)[red]{};
				
				\coordinate (i01) at (3,-1);
				\coordinate (o01) at (3,1);
				\coordinate (i11) at (6, 0);
				\coordinate (o11) at (6,2);
				
				\node at (3.5,-0.5)[regular polygon,regular polygon sides=3,draw,fill=black!10,inner sep=1pt] (n01) {$n_0$};
				\node at (5.5,1.3)[regular polygon,regular polygon sides=3,draw,fill=black!10,inner sep=1pt] (n11) {$n_{1}$};
				\node at (4.5,0.5)[regular polygon,regular polygon sides=3,shape border rotate=45,draw=red,fill=black!10,inner sep=1pt] (n21) {$n_{2}$};
				
				\draw[->] (i01.north)-- (n01.south){};
				\draw[->] (n01.north)-- (o01.south){};
				\draw[->] (i11.north)-- (n11.south){};
				\draw[->] (n11.north)-- (o11.south){};
				\draw[->] (n01.north)-- (n21)[red]{};
				\draw[->] (n21.east)-- (n11.south)[red]{};
				
				\begin{scope}[on background layer]
				
				\node[fit= (i0)(o1),rectangle,fill=black!15, rounded corners=0.5cm,blur shadow={shadow blur steps=5}] (pre){};

				\end{scope}
				
				\begin{scope}[on background layer]
				
				\node[fit= (i01)(o11),rectangle,fill=black!15, rounded corners=0.5cm,blur shadow={shadow blur steps=5}] (post){};

				\end{scope}
				
				\draw[->] (pre)-- (post)[ultra thick]{};
				
				\end{tikzpicture}}
			\caption{insert\_network}
			\label{fig:div_conn}
		\end{subfigure}
		\qquad
		\begin{subfigure}[b]{0.22\textwidth}
			\resizebox {3.8cm} {!}{
				\begin{tikzpicture}
				
				\coordinate (i0) at (-1,-1);
				\coordinate (o0) at (-1,1);
				\coordinate (i1) at (2, 0);
				\coordinate (o1) at (2,2);
				
				\node at (-0.5,-0.5)[regular polygon,regular polygon sides=3,draw,fill=black!10,inner sep=1pt] (n0) {$n_0$};
				\node at (1.5,1.3)[regular polygon,regular polygon sides=3,draw,fill=black!10,inner sep=1pt] (n1) {$n_{1}$};
				\node at (0.5,0.5)[regular polygon,regular polygon sides=3,shape border rotate=45,draw=red,fill=black!10,inner sep=1pt] (n2) {$n_{2}$};
				
				\draw[->] (i0.north)-- (n0.south){};
				\draw[->] (n0.north)-- (o0.south){};
				\draw[->] (i1.north)-- (n1.south){};
				\draw[->] (n1.north)-- (o1.south){};
				\draw[->] (n0.north)-- (n2)[red]{};
				\draw[->] (n2.east)-- (n1.south)[red]{};
				
				\coordinate (i01) at (3,-1);
				\coordinate (o01) at (3,1);
				\coordinate (i11) at (6, 0);
				\coordinate (o11) at (6,2);
				
				\node at (3.5,-0.5)[regular polygon,regular polygon sides=3,draw,fill=black!10,inner sep=1pt] (n01) {$n_0$};
				\node at (5.5,1.3)[regular polygon,regular polygon sides=3,draw,fill=black!10,inner sep=1pt] (n11) {$n_{1}$};

				\draw[->] (i01.north)-- (n01.south){};
				\draw[->] (n01.north)-- (o01.south){};
				\draw[->] (i11.north)-- (n11.south){};
				\draw[->] (n11.north)-- (o11.south){};
				\draw[->] (n01.north)-- (n11.south)[red]{};
				
				
				\begin{scope}[on background layer]
				
				\node[fit= (i0)(o1),rectangle,fill=black!15, rounded corners=0.5cm,blur shadow={shadow blur steps=5}] (pre){};

				\end{scope}
				
				\begin{scope}[on background layer]
				
				\node[fit= (i01)(o11),rectangle,fill=black!15, rounded corners=0.5cm,blur shadow={shadow blur steps=5}] (post){};

				\end{scope}
				
				\draw[->] (pre)-- (post)[ultra thick]{};
				
				\end{tikzpicture}}
			
			\caption{delete\_network}
			\label{fig:byp_net}
		\end{subfigure}
		\qquad
		\begin{subfigure}[b]{0.22\textwidth}
			\resizebox {3.8cm} {!}{
				\begin{tikzpicture}
				
				\coordinate (i0) at (-1,-1);
				\coordinate (o0) at (-1,1);
				\coordinate (i1) at (2, 0);
				\coordinate (o1) at (2,2);
				
				\node at (-0.5,-0.5)[regular polygon,regular polygon sides=3,draw,fill=black!10,inner sep=1pt] (n0) {$n_0$};
				\node at (1.5,1.3)[regular polygon,regular polygon sides=3,draw,fill=black!10,inner sep=1pt] (n1) {$n_{1}$};
				\node at (0.5,0.5)[regular polygon,regular polygon sides=3,shape border rotate=45,draw=red,fill=black!10,inner sep=1pt] (n2) {$n_{2}$};
				
				\draw[->] (i0.north)-- (n0.south){};
				\draw[->] (n0.north)-- (o0.south){};
				\draw[->] (i1.north)-- (n1.south){};
				\draw[->] (n1.north)-- (o1.south){};
				\draw[->] (n0.north)-- (n2)[red]{};
				\draw[->] (n2.east)-- (n1.south)[red]{};
				
				\coordinate (i01) at (3,-1);
				\coordinate (o01) at (3,1);
				\coordinate (i11) at (6, 0);
				\coordinate (o11) at (6,2);
				
				\node at (3.5,-0.5)[regular polygon,regular polygon sides=3,draw,fill=black!10,inner sep=1pt] (n01) {$n_0$};
				\node at (5.5,1.3)[regular polygon,regular polygon sides=3,draw,fill=black!10,inner sep=1pt] (n11) {$n_{1}$};
				\node at (4.8,-0.2)[regular polygon,regular polygon sides=3,shape border rotate=45,draw=red,fill=black!10,inner sep=1pt] (n21) {$n_{2}$};
				\node at (4,0.7)[regular polygon,regular polygon sides=3,shape border rotate=45,draw=red,fill=black!10,inner sep=1pt] (n3) {$n_{3}$};
				
				\draw[->] (i01.north)-- (n01.south){};
				\draw[->] (n01.north)-- (o01.south){};
				
				\draw[->] (n11.north)-- (o11.south){};
				\draw[->] (n01.north)-- (n21.west)[red]{};
				\draw[->] (n01.north)-- (n3.west)[red]{};
				
				\draw[->] (n21.east)-- (n11.south)[red]{};
				\draw[->] (n3.east)-- (n11.south)[red]{};
				
				
				\begin{scope}[on background layer]
				
				\node[fit= (i0)(o1),rectangle,fill=black!15, rounded corners=0.5cm,blur shadow={shadow blur steps=5}] (pre){};

				\end{scope}
				
				\begin{scope}[on background layer]
				
				\node[fit= (i01)(o11),rectangle,fill=black!15, rounded corners=0.5cm,blur shadow={shadow blur steps=5}] (post){};

				\end{scope}
				
				\draw[->] (pre)-- (post)[ultra thick]{};
				
				\end{tikzpicture}}
			
			\caption{clone\_network}
			\label{fig:dup_net}
		\end{subfigure}
		
		\caption{Examples of the different operators. In all cases, the variation is performed relative to the VALP sub-network in the middle of the figure (in red). For Figures \ref{fig:add_conn}, \ref{fig:delete_conn}, and \ref{fig:div_conn}, a connection. For Figures \ref{fig:byp_net}, and \ref{fig:dup_net}, a network ($n_2$).}
		\label{fig:operators}
	\end{center}
\end{figure}

Graphical examples of how these operators work are shown in Fig.~\ref{fig:operators}.

\subsection{Hyperparameters}

Searching for the optimal model architecture (the sub-networks and how they are interconnected within the model) would only raise the model to a certain point, as the rest of the model components need to be synchronized to obtain an optimal performance. This is the case of the loss functions used to optimize the weights of the neural model and other hyperparameters, as the SGD algorithm. Other aspects related to training, such as the learning rate and batch size, also have to be properly set. With this in mind, we define the following variation operators, all of which are gentle:

\begin{itemize}
	\item \texttt{change\_lr} changes the learning rate of a model output. For example, if convergence is detected in the \emph{Historic sub-loss function information}, its learning rate can be decreased, aiming at improving the effectiveness of the training procedure over that specific objective. 
	\item \texttt{change\_sgd} changes the SGD algorithm used to optimize the weights of the model with respect to a model output.
	\item \texttt{change\_bs} changes the size of the batch used at each training epoch.
\end{itemize}

\subsection{Crossover operator}

In this multi-objective scenario, the objectives are independent of each other to some degree, as each output will normally have some exclusive sub-DNNs (and therefore, weights). Employing crossover-like operators enables parts of models to be cherry-picked for constructing other models with the \textit{best parts} of each one. We define a crossover operator based on the donation between models: 


\begin{itemize}
	\item Exclusive subgraph crossover: This aggressive operator can be applied when, based on the Relative Performance measure, a model behaving adequately in multiple tasks fails at another one. A model with a top performance in that last task is selected as the donor of the exclusive subgraph of that output for the first model, the host, which has its exclusive subgraph replaced by the donation.
\end{itemize}

\section{Experiments}\label{sec:experiments}

We have designed a set of experiments in order to validate some of the general guidelines for the NAS framework proposed in this paper.

Several works have reported that starting from \textit{simple} neural models with relatively few parameters and allowing them to evolve towards more complex structures yields positive results \cite{lu_compnet_2018}. The experiments described in this section consist of the employment of the proposed search guidelines with this same mindset. We consider a model with a number of components close to the minimum (roughly one sub-network per model output) to provide the required output to be on its initial stages, whereas a mature model would consist of a more complex structure with more sub-networks and connections.

\subsection{Test Benchmark}

For the different parts of the experimentation, we have built two artificial and similar problems. Both of them consist of extensions of two widely known problems, MNIST \cite{lecun_mnist_2010} and Fashion MNIST \cite{xiao_fashion-mnist:_2017}. The two datasets are composed of images of the same number of pixels (784) arranged in the same manner (28$\times$ 28), although the former is composed of handwritten digits, and the latter consists of pictures of clothing pieces. We define the multiobjective version of both problems \cite{garciarena_expanding_2018}. 

\begin{itemize}
	\item \textbf{Classification objective:} This is the original definition of the problem. It consists of correctly classifying the observations in the dataset into one of the 10 possible classes.
	\item \textbf{Histogram prediction objective:} This objective consists of correctly predicting the histogram of the pixels in the images. The images being grayscale, a single histogram (of 8 bins) can be computed and associated with each one of the examples.
	\item \textbf{Image sampling objective:} This last objective consists of sampling images similar to those in the dataset.
\end{itemize}

This way, we can test the performance of the operators when acting in an environment where the outputs are related to a single data input.

We define the two separate problems to simulate the scenario in which the rules are inferred from one set of experiments, and are then applied to another, more complex problem. The two problems having very similar characteristics in terms of the number of examples and features as well as data type and the number of classes is purely coincidental, as this approach could be tested in problems of varying data inputs, outputs, and characteristics of both.

\subsection{Initial experimentation} \label{sec:preliminarydesc}

The first step consists of testing the proposed metrics and operators isolated from the NAS framework. This way, we will be able to extract valuable information about how to use the information given by the metrics with the final goal of deciding which operator and where it should be applied in a NAS process.

In order to assess the impact that each operator can have in different scenarios (these being described by the values obtained from the different metrics), we perform an exploratory search over the space of \textit{medium-sized} VALPs (i.e., twice as many sub-networks as model outputs). 
In this experimentation, we will be able to observe the difference between applying gentle mutation operators over their aggressive counterparts.

Additionally, and this is the main goal of this experimental section, we aim at setting the grounds of the set of rules which will be helpful to improve the efficiency of future NAS runs. To that end, we attempt to identify which operators offer the largest improvement potential. Because the rules we are looking for should not be tied to the particular problem used in this instance, we are relying on the metrics defined in Section~\ref{sec:metrics} instead of the common metrics for assessing the performance of a prediction model (e.g., accuracy for a classification model).

Choosing the MNIST problem, we test the effect of the mutation operators defined in Section~\ref{sec:classicMut}. To that end

\begin{enumerate}
	\item 100 VALPs are randomly created (as described in the supplementary part of this work and in \cite{garciarena_towards_2021}) and trained for $\sim$67 epochs (20,000 batches of size 200).
	\item Every operator is applied to different clones of each VALP. The operators are applied to each component of the VALPs only if structural correctness is guaranteed.
	\item Every VALP is retrained to adjust the weights of the model to the variation for $\sim$17 additional epochs (5,000 more batches).
\end{enumerate}\

To determine the quality of the VALPs at each point, we have evaluated them before the modification and after the secondary training.

\subsection{Main experimentation}

In this second step, we want to employ the knowledge obtained in the first step on a NAS procedure. With that goal in mind, we propose a common HC algorithm (Figure~\ref{alg:HC} contains a pseudo-code form of the method) with two different implementations: the common approach, in which operators are chosen randomly, and the smart approach, in which the most promising operator is chosen. The pseudo-code makes use of the following functions:

\begin{itemize}
	\item \textbf{random\_VALP()}: This function randomly initializes a VALP, with a limited number of components.
	\item \textbf{evaluate($model$)}: Given a VALP, this function evaluates the model and returns one value per model output. In this work, it consists of a triple, since the problem has three objectives.
	\item \textbf{select\_operator($model$)}: Given a VALP, this function selects the operator to be applied. The difference between the random and the \textit{smart} HC approach resides in the implementation of this function.
	\item \textbf{variation($model, op$)}: Given a VALP and an operator, this function generates a neighbor of the VALP by applying the operator.
	\item $\boldmath{<}$: This operator compares two tuples of values. In this case, if at least two of the three values of the operand on the left are lower than their corresponding values on the right, it returns True. Otherwise, False is returned.
	
\end{itemize}

\begin{figure}
	\centering
	\begin{algorithmic}[1]
		\Procedure{StochasticHC}{$step\_limit$}
		\State $current\gets$ random\_VALP()
		\State $curr\_fitness\gets$ evaluate($current$)
		\State $step\gets 0$
		\While{$step<step\_limit$}
		\State $op\gets$ select\_operator($current$)
		\State $candidate\gets$ variation($current, op$)
		\State $cand\_fitness\gets$ evaluate($candidate$)
		\If{$cand\_fitness < curr\_fitness$}
		\State $current\gets candidate$
		\State $curr\_fitness\gets cand\_fitness$
		\EndIf
		\State $step\gets step + 1$
		\EndWhile
		\State \textbf{return} $current, curr\_fitness$
		\EndProcedure
	\end{algorithmic}
	\caption{HC approach used in the experiments.}
	\label{alg:HC}
\end{figure}

The two variants of the algorithm are tested on the (slightly more difficult) Fashion-MNIST multiobjective problem. Following the start-simple-and-sophisticate approach, a random solution is initialized containing between one and two times as many sub-networks as model outputs. Then, both variations of the algorithm are applied to search for VALP structures.

The VALP configurations used as the starting point are trained for 5.000 batches. At each step of the HC algorithm, the modified model is retrained for 1.000 additional batches. Each HC is run $30$ times with different random seeds in order to avoid possible bias, product of the stochastic component of the method. $60$ steps are awarded to each search method.

As in the initial experiment, we constrain the set of variation operators to be investigated to those defined in Section~\ref{sec:classicMut}.

\section{Results}\label{sec:results}

\subsection{Initial experimentation}

\begin{figure}
	\begin{center}
		\includegraphics[width=0.4\paperwidth]{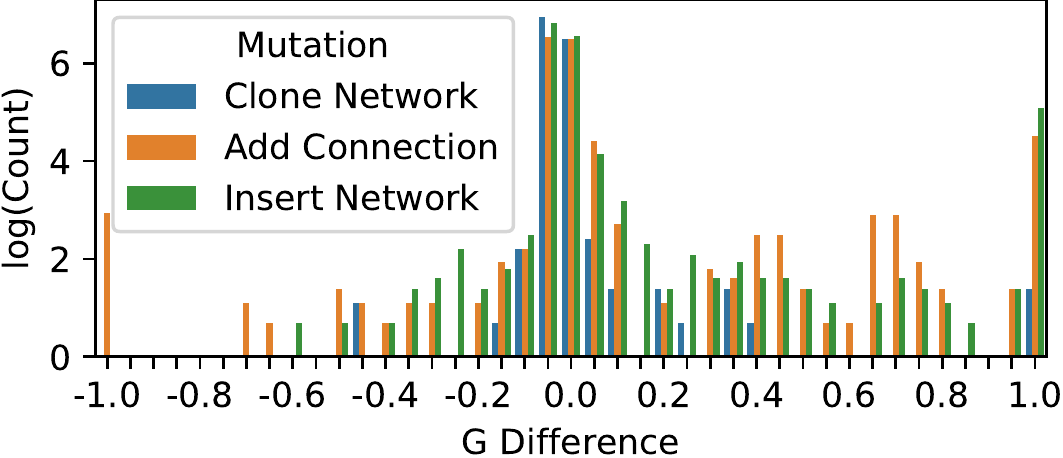}
		\caption{$G$ difference observed when comparing the improvements obtained by VALPs after being modified by gentle operators, and their aggressive versions. Positive values indicate a better performance of the gentle operator, whereas negative ones do the exact opposite. Larger numbers represent larger differences between operators.}
		\label{fig:differences}
	\end{center}
\end{figure}

First, we want to investigate whether the gentle operators consistently perform better than their aggressive counterparts. To that end, we have computed the improvement observed in the VALPs between the end of the first training session and after it has been modified and retrained. With the improvement measured -performance after second training divided by performance after the first training, both measurements in logarithmic scale- we subtract the improvement observed due to the application of gentle operators to the improvement caused by their corresponding aggressive counterparts. This metric $G$ serves as a measure of the gain or advantage of using one class of operator over the other. Figure~\ref{fig:differences} shows the frequency (y axis) of the $G$ difference values (x axis). The more positive they are, the bigger the difference in favor of the gentle operator. Any difference superior to one is cut to that value to improve the visualization of the figure.

As can be observed in Figure~\ref{fig:differences}, the gentle operators have outperformed the aggressive ones considerably more frequently than the other way around, especially taking into account extreme differences (values over 1). Gentle operators are, in general, conservative variations when it comes to increasing or decreasing the performance of the model. In closer comparisons, the gentle operators also tend to produce a bigger improvement in model quality. However, although less frequent, there are many cases in which the aggressive operators had a more positive impact than their gentle counterparts. The \textit{noise} these operators introduce into the model, in the form of random weights, appears to be able to \textit{shake} the model from a local optimum, from which the gentle operator could not make it escape. This is especially visible in the cases in which extreme improvements were achieved by the aggressive operators.

The presence of these last cases suggests that the employment of aggressive operators is not only viable, but advisable in some scenarios. This theory is also backed up by statistical testing. After the null hypothesis of all mutations producing the same effects was rejected by the Kruskal-Wallis statistical test \cite{kruskal_use_1952}, the Dunn post-hoc test \cite{dunn_multiple_1964} found significant differences between all pairwise comparisons between mutations -p-value $<$ 0.0007- except for one, the comparison between the aggressive and gentle version of the connection adding operator. This is probably due to the high number of extreme differences in improvements. In other cases, although many of the measurements fall close to zero, there is a significantly larger amount of differences on the positive side as opposed to the negative side, including a substantial amount of extreme values over one. This explains the significant differences corroborated by the test.

We now address the question of how to create the set of rules which helps NAS algorithms to correctly identify the \textit{best operator} given one model.

Regarding the second part of the experimentation, we attempt to define the metrics that will eventually guide future NAS runs. With that goal in mind, we aim at observing, given the metric values (from those defined in Section~\ref{sec:metrics}), which operator(s) produced the largest gains. The two metrics which have produced the most significant differences among the analyzed mutation operators were the historic sub-loss information and the module intervention.

\begin{figure*}
	\begin{center}
		\begin{subfigure}[b]{0.32\textwidth}
			\includegraphics [width=\textwidth]{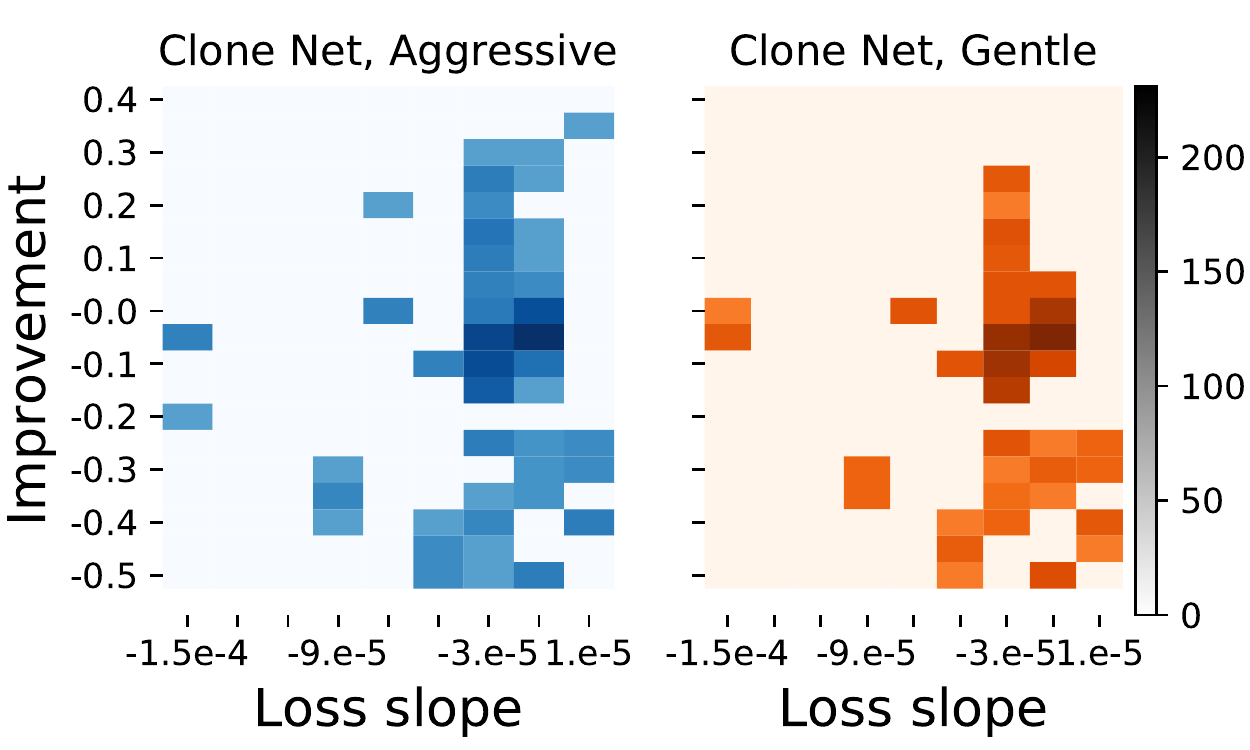}
			\caption{Clone Net mutation operator.}
			\label{fig:clone_loss}
		\end{subfigure}
		\begin{subfigure}[b]{0.32\textwidth}
			\includegraphics [width=\textwidth]{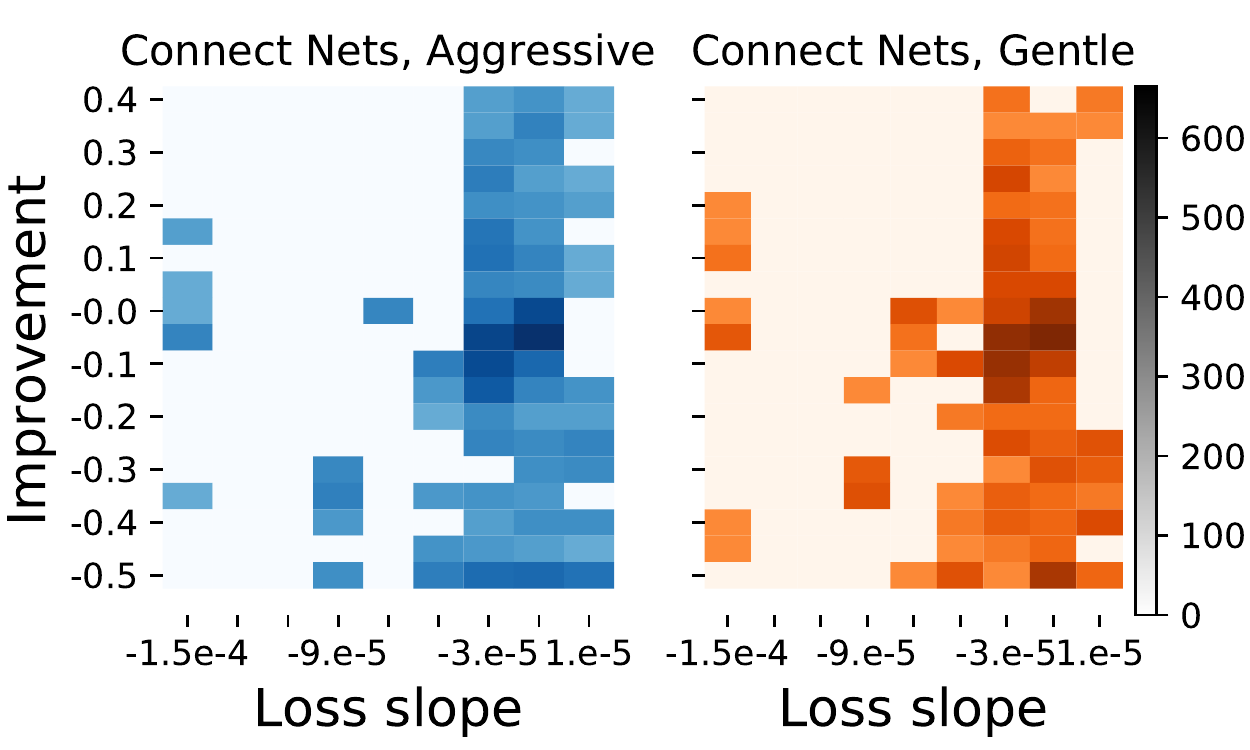}
			\caption{Add Connection mutation operator.}
			\label{fig:connect_loss}
		\end{subfigure}
		\begin{subfigure}[b]{0.32\textwidth}
			\includegraphics [width=\textwidth]{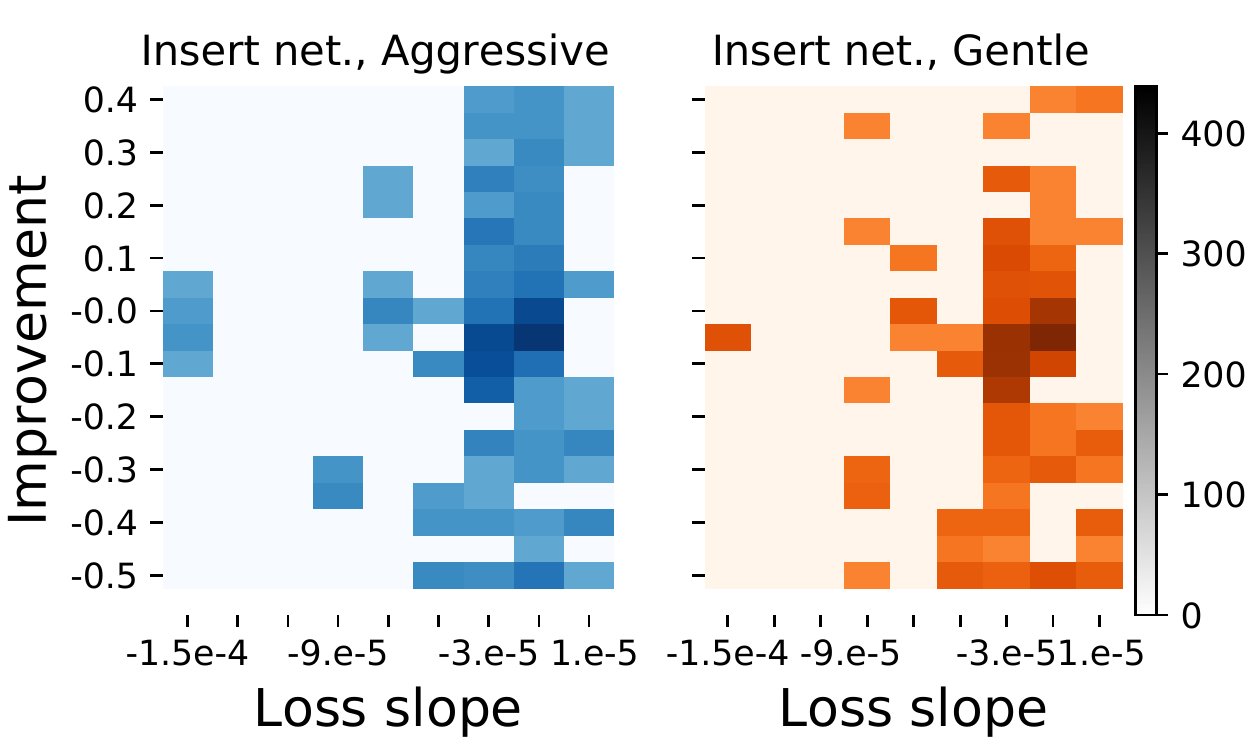}
			\caption{Insert network mutation operator.}
			\label{fig:div_loss}
		\end{subfigure}
		\caption{Percentage improvement observed over the regression output of the VALP (in the y axis, the lower the value, the larger improvement) in logarithmic scale, by different mutation operators, with respect to the slope of the historic sub-loss function evolution (x axis). The color darkness represents the number of mutations that registered the improvement in the y axis. The subfigures on the left-hand side represent data relative to the aggressive version of the operators, whereas the ones on the right-hand side show information about the gentle ones.}
		\label{fig:mut_loss}
	\end{center}
\end{figure*}

Regarding the loss function slope, Figure~\ref{fig:mut_loss} shows the percentage of improvement observed in a VALP after it was modified by the gentle operators and their aggressive counterpart (in the y axis, in logarithmic scale), regarding the historic sub-loss function of the regression output of the VALP (x axis). The improvement percentages (the lower, the larger improvement) have been cut to the $[-0.5, 0.5]$ range. The loss slope also only considers a minimum value of $-0.00015$. As can be seen in the figure, most improvements are marginal, just below the 0 mark. However, note that the goal of this analysis is to observe in which cases the operators can produce a significant improvement to the search, rather than how often they are able or unable to do so. The random application of an operator, as can be seen in the distributions in Figure~\ref{fig:mut_loss}, would very unlikely result in a drastic improvement. Because of this, some interesting insights can be extracted from this figure.

For example, because of the lack of existence of large performance decline when applying the clone net operator (Figure~\ref{fig:clone_loss}) if the loss function is still decreasing (left-hand side of the figures), we can conclude that this is usually a beneficial mutation. The connection adding operator (Figure~\ref{fig:connect_loss}) was also able to produce large performance gains when the loss function of an output is still steeply decreasing. This means that these kinds of changes are beneficial, especially the gentle form, when the loss function is still decreasing. 

The mutation that places a network in the middle of a connection (Figure~\ref{fig:div_loss}) was able to produce significant changes (both improvement and deterioration) when the slope of the loss function is smaller. This is especially true for the aggressive version, as a significant number of drastic improvements was observed compared to the gentle version. This means that applying it on an output which has saddled in a poor local optimum can make the model to dramatically improve, while a drastic performance loss would not hurt the search, as the local optima was not desirable anyway. 

A similar set of figures has been generated for the network relevance metric. Relevance consists of the change observed between the two stages of the model, before and after being affected by the module intervention approach, and it is also measured in percentage points. This way, if no change was observed in a model output after being affected, a $1$ is recorded. If the performance was halved (e.g., only half of the observations which were previously correctly classified are correctly classified after modifying the model), a $2$ is recorded. 

\begin{figure*}
	\begin{center}
		\begin{subfigure}[b]{0.32\textwidth}
			\includegraphics [width=\textwidth]{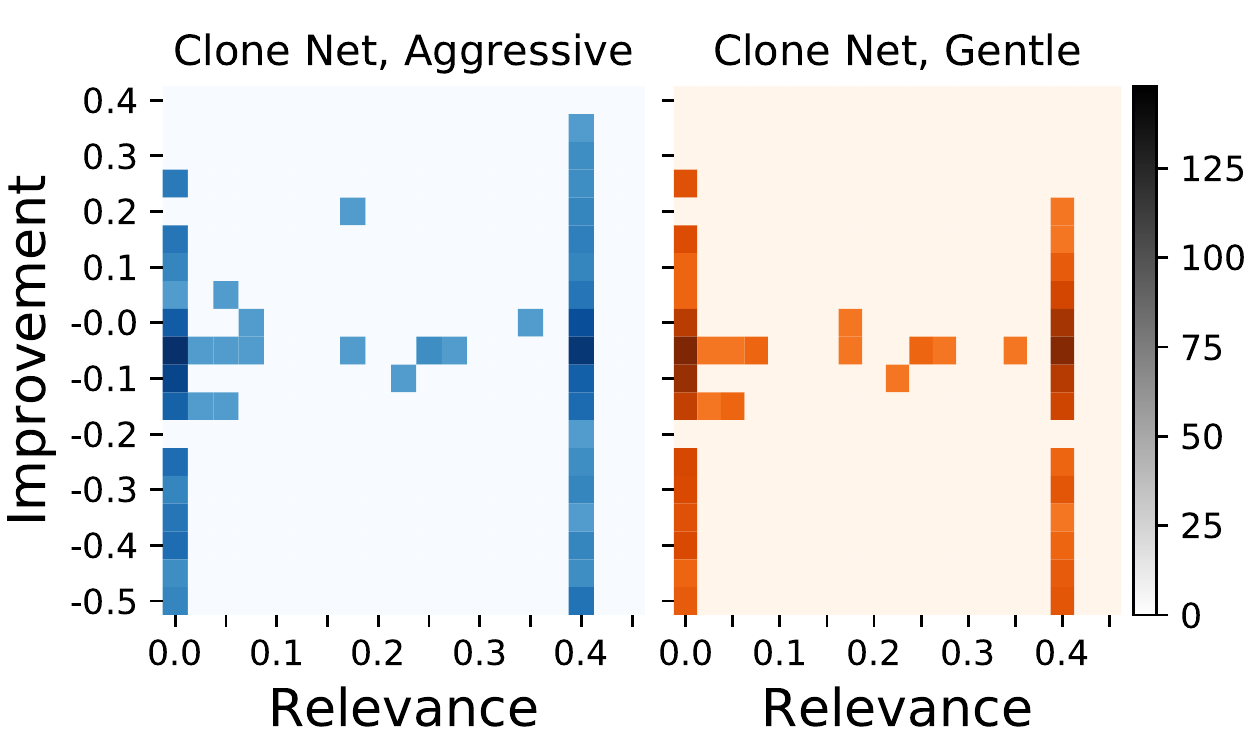}
			\caption{Clone Net mutation operator.}
			\label{fig:clone_rel}
		\end{subfigure}
		\begin{subfigure}[b]{0.32\textwidth}
			\includegraphics [width=\textwidth]{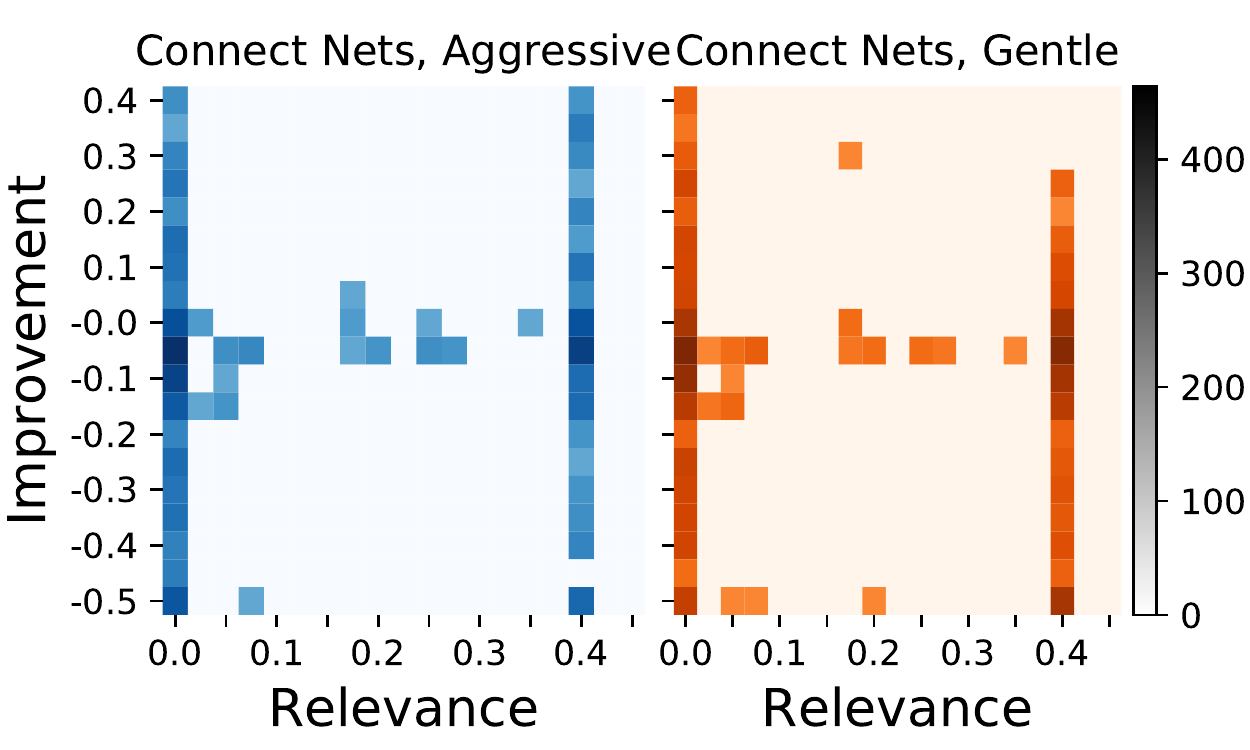}
			\caption{Add Connection mutation operator.}
			\label{fig:connect_rel}
		\end{subfigure}
		\begin{subfigure}[b]{0.32\textwidth}
			\includegraphics [width=\textwidth]{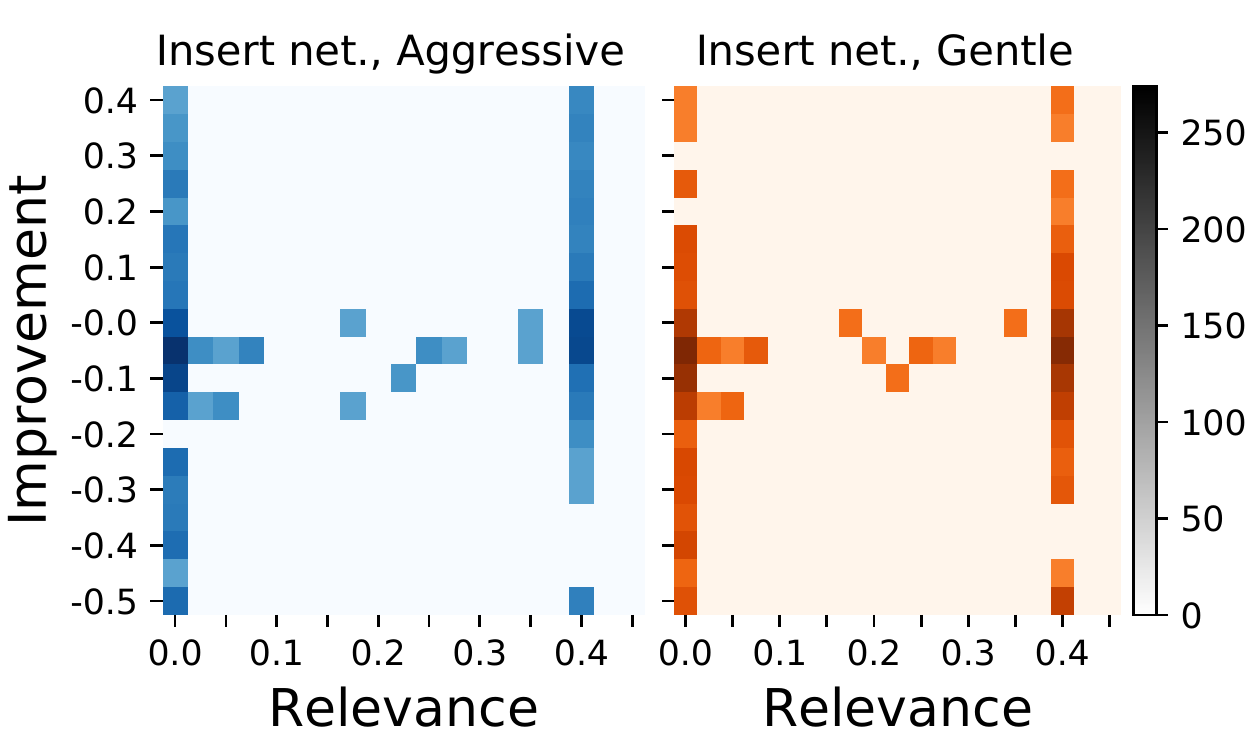}
			\caption{Insert network mutation operator.}
			\label{fig:div_rel}
		\end{subfigure}
		\caption{Percentage improvement observed over the regression output of the VALP in logarithmic scale (in the y axis), by different mutation operators, with respect to the relevance of the network affected by the operator, also in logarithmic scale (x axis, the larger, the more relevant a network to the output).}
		\label{fig:mut_rel}
	\end{center}
\end{figure*}

In the case of Figure~\ref{fig:mut_rel}, because the performance of a regression output can decrease indefinitely, the relevance has been cut to $0.4$ (in logarithmic scale). As can be observed in the top right corner of the figures, when modifying a network relevant to an output, the result, as expected, can be very bad if the mutation is an aggressive one. The gentle network cloning appears to be a conservative choice when it comes to a relevant net, given the few cases in which performance declines have been observed. A similar effect can be observed with the insert network operator, as it produced more improvements when applied to relevant networks. The connection adding operator in this case is not advisable with relevant networks.

\subsection{Operator per network characterization}\label{sec:rules}

With the insights made in the previous section, we have defined the following set of rules to display the potential of this kind of guided searches.

\begin{itemize}
	\item When a network involved in a loss function registers a steep descent, the gentle version of the network cloning mutation can be applied.
	\item If a network is part of an output which is moderately descending, the add connection operator can be applied.
	\item When a network is part of an output in a local optimum, the insert network can be applied.
	\item If a network is not relevant for some outputs it is connected to, but is for other ones, the delete connection, the insert network, or the aggressive version of the clone network can be applied.
\end{itemize}

These rules, along with these additional ones,

\begin{itemize}
	\item When a network is not relevant for an output it is connected to, the delete connection operator can be applied.
	\item If a network is not relevant for any output, the network deletion operator can be applied.
\end{itemize}

have been compiled into the mutation selection guidelines, which are going to be used in the HC algorithm. 

A more \textit{sophisticated} usage of the metrics and variation operator characterization could be carried out. Defining a metamodel that is able to capture the patterns between model states and operator application with the largest benefit for an efficient search of the DNN structure could improve the obtained results \cite{garciarena_exploitation_2021}.

\subsection{Main experimentation}

The threshold values for determining whether a loss function is descending or not, or how relevant a sub-network is, are parameters of the NAS algorithm. In this case, they are estimated from the initial experimentation. A loss slope larger than $-10^{-10}$ is considered to be stuck, and if smaller than $-2\times 10^{-5}$, it is determined to be steeply descending. Anything in between these two values is considered to be moderately descending. A network is considered to be relevant to an output if the performance of the VALP in that output decreases by $20\%$ or less when it is intervened.

These values could be used as a reference for setting these threshold values in the future, taking into account the magnitudes of the problems being dealt with in each case.

\subsection{Operator selection}

With these defined criteria, all networks within a VALP can be modified by several operators at each stage. Therefore, we define a hierarchy in which the operators are organized according to the priority they are given to modify the models. 

\begin{enumerate}
	\item Reducers: Because we pursue efficient models, any network or connection which is not valuable for the overall performance should be deleted.
	\item Aggressive expanders: Any network which, according to the rules defined in the previous section, can be affected by an aggressive expander operator, is assumed not to be working properly, and this is the second priority.
	\item Gentle expanders: Giving more modeling power to a model only makes sense when all its resources are being effectively used, and therefore this is the last type of operators to be taken into account.
\end{enumerate}

\begin{figure*}
	\begin{center}
		\includegraphics[trim=0 0 16 0,clip,width=0.75\textwidth]{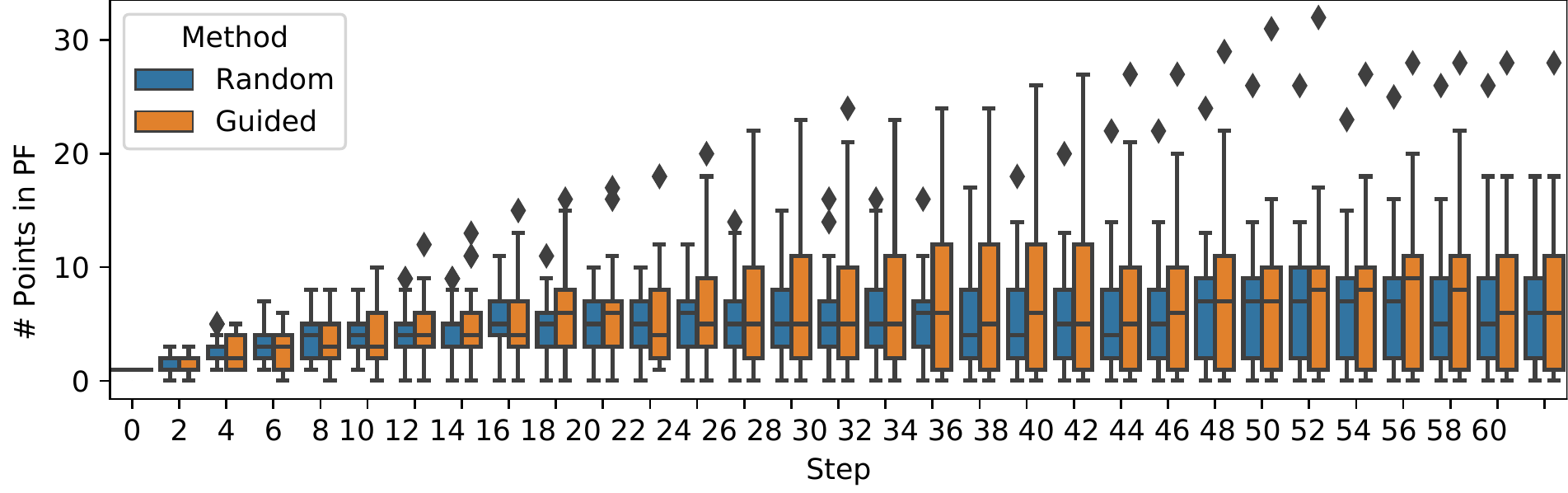}
		\caption{Boxplot showing the number of points (y axis) in the PFs generated from combining all HC runs (blue for random, orange for guided), per step (x axis). The larger the number of points in the PF, the better the performance of the algorithm.}
		\label{fig:paretos}
	\end{center}
\end{figure*}

At the time of selecting the operator to be applied, within the set of operators with the highest priority, one is chosen at random. When a selected operator is not able to improve the current model, it is not included among the candidate operators to generate a neighbor of the model in the next step. If no operator from those selected by this method is able to create a candidate model that improves the current solution, a random gentle operator is applied to a random network in the model.

Since we deal with multi-task problems where multiple objectives have to be simultaneously optimized, the question of deciding what model is the better one is not trivial, and, therefore, neither is when comparing search algorithms. We thus resort to using two Pareto front-based approaches to compare the quality of the VALP structures found. Further analysis of these results can be found as part of the supplementary work.

For the first comparison, we take each of the 30 pairs of runs separately, considering as pairs those runs which start from the same random VALP structure and use the random and guided HC approaches. At each step, all the structures found (across the whole search) by a pair of runs are compiled into a single set, and a Pareto front is computed, considering the three outputs of the model. This way, in, for example, the fourth step, we have 30 different PFs, each being composed of at most eight points, four from each of the corresponding runs (one per completed step). Next, the points in the PF from each HC approach are counted. In Figure~\ref{fig:paretos}, boxplots are presented, which display the number of points in the PFs (y axis) by each approach (orange for the guided HC and blue for the random version), in each step (x axis).

As can be seen in Figure~\ref{fig:paretos}, in the initial 20 steps, both versions of the algorithm work similarly, with a slight advantage for the random HC. This trend changes after the 20th step, where, although the median remains similar, the top results are clearly produced by the guided version of the algorithm.

Interestingly, both the random and the guided versions have produced one run each which generates a number of points in their corresponding PFs far superior to the rest. These outliers are also higher in the guided version.

\begin{figure*}
	\begin{center}
		\includegraphics[width=0.75\textwidth]{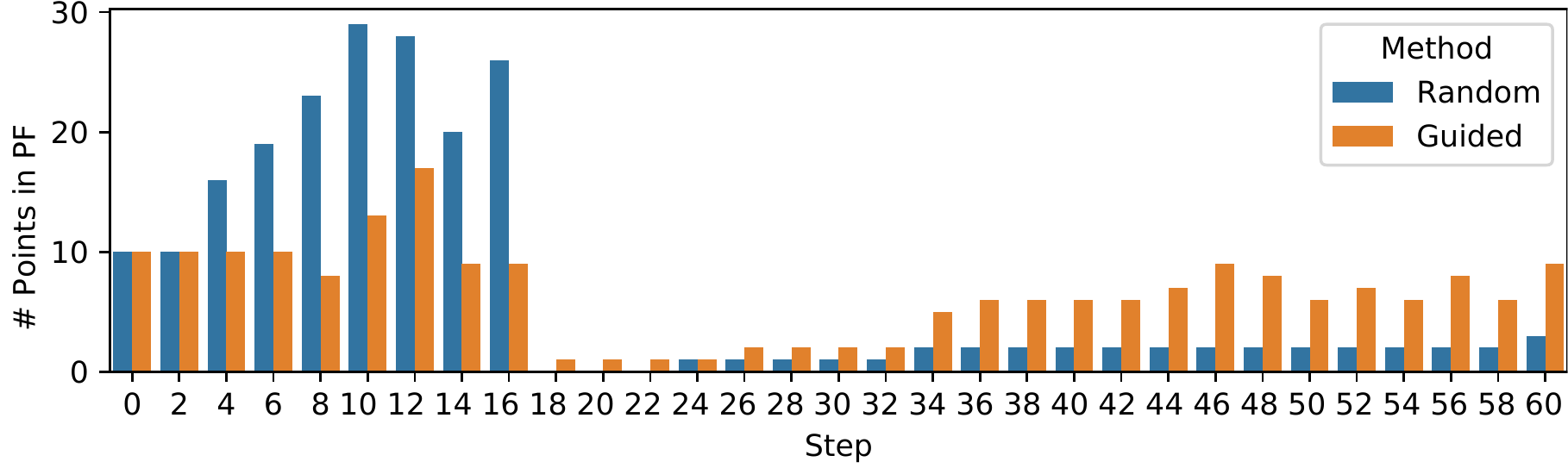}
		\caption{Barplot showing the number of points (y axis) in the combined PF from each approach (blue for random, orange for guided), per step (x axis). The larger the number of points in the PF, the better the performance of the algorithm.}
		\label{fig:crossparetos}
	\end{center}
\end{figure*}

Secondly, we consider all 30 runs together, in order to know what algorithm is able to obtain the best results, overall. In this case, instead of constructing one PF per step and pair of runs, we simply construct a single PF from all the points found across runs limited only by the step. Again, all the found structures until a step are considered in each step. The results are shown in Figure~\ref{fig:crossparetos}.

Although Figures \ref{fig:paretos} and \ref{fig:crossparetos} look dissimilar, the information shown coincides. During the initial stages of the search (the initial 16-17 steps), the algorithms are searching for the best area to exploit, at which the random HC seems to outperform the guided version. This comes as no surprise, as the randomized approach does not focus on a \textit{search path} to follow. Because it can perform modifications in any place within the model structure, the model can improve or lose performance continuously in different outputs. This helps a larger presence of points generated by the random HC in the PFs shown in Figure~\ref{fig:crossparetos}, as opposed to the guided version, which focuses on improving certain aspects of the model -the efficiency of the sub-networks- before starting to seek performance improvements. That first phase ends near the 18th step, as one of the guided runs achieves one VALP configuration capable of dominating all the ones found during all searches. Slowly, other points start to form the PF, most of which belong to the guided runs. This shows the benefit of the guided search over the randomized one in the long term when performing intelligently chosen moves.



\section{Conclusions}\label{sec:conclusions}

This work is framed in the neural architecture search field. Our efforts are focused on compiling a set of guidelines which aims at maximizing the effectiveness of the application of variation operators to model structures during a structural  search procedure, illustrated using a complex scenario, HMTL. More specifically, we first identify several metrics which can be used to determine the level of importance of different sub-DNNs in the overall performance. Secondly, we compile a set of variation operators previously used in NAS procedures described in the literature and classify them according to the effect they have on the complexity and performance of the model.
Next, we conduct an extensive exploratory search on how these operators affect the performance of a \textit{medium-sized} model, in order to identify patterns that relate the defined metrics and the improvements in the models. These patterns are latter transformed into a set of guidelines for enhancing the efficiency of future NAS searches. These guidelines add one level of sophistication to current NAS algorithms, as, opposed to the common practice of randomly selecting a variation operator, a more informed choice is made, which can save the need to evaluate DNN structures affected by the \textit{wrong} operator. Finally, a comparison of the performance of the two variants of the NAS search -blind versus guided by the introduced rules- is presented as an illustration of the gains that could be obtained in NAS efficiency.

The main contribution of this work is the methodology for diagnosing the state of an HMTL model and identifying the relevance of its different components, and the application of these metrics for more efficient NAS algorithms. One key for this goal is the set of metrics defined with this purpose, although others which complement those introduced in this work could result in more valuable information about the model, ultimately making the processes more efficient.

The experiments conducted in this work serve as a blueprint for implementing the presented ideas to other problems and domains, as they have already served the purpose of efficiently exploring a complex search space using. Although the conducted NAS runs can be considered as \textit{simple}, the defined methodology is not restrained to be applied in such scenarios.

The study and application of these (or similar) methodologies to other problem definitions; e.g., NAS types (e.g., neuroevolutionary algorithms), or DNN types (single objective, convolutional DNNs, etc.); is left as future work, as are the employment of some of the metrics and operators described in this work, which were not tested in the experiments; and studying the possibility of adaptation of reinforcement learning and gradient based NAS approaches to the HMTL framework.

\section*{Acknowledgment}

This work has received support from the KK-2020/00049 (3KIA through the ELKARTEK program) PID2019-104966GB-I00 (Spanish Ministry of Science and Innovation) and IT-1244-19 (Basque Government) programs. Unai Garciarena holds a predoctoral grant (PIF16/238) by the University of the Basque Country. We gratefully acknowledge the support of NVIDIA Corporation with the donation of a Titan X Pascal GPU used to accelerate the process of training the models used in this work.


\ifCLASSOPTIONcaptionsoff
  \newpage
\fi



%
\bibliographystyle{abbrv}
\bibliography{references.bib}

%

\begin{IEEEbiographynophoto}{Unai Garciarena}
received his Ph.D. degree in computer science in 2021, in the University of the Basque Country (UPV/EHU). His main research interests are generative modeling, supervised classification, and optimization.
\end{IEEEbiographynophoto}

\begin{IEEEbiographynophoto}{Roberto Santana}
received a Ph.D. degree in Mathematics from the University of Havana, Havana, Cuba, in 2005, and a Ph.D. degree in Computer Science from the University of the Basque Country, Spain, in 2006, where he is a researcher. His research interests include machine learning, evolutionary computation, probabilistic graphical models, and neuroscience.
\end{IEEEbiographynophoto}


\begin{IEEEbiographynophoto}{Alexander Mendiburu}
received the Ph.D. degree from the University of the Basque Country, Spain, in 2006. Since 1999, he has been a Lecturer with the Department of Computer Architecture and Technology, University of the Basque Country. His current research interests include evolutionary computation, probabilistic graphical models, and parallel computing.
\end{IEEEbiographynophoto}




\end{document}